\documentclass[conference]{IEEEtran}
\IEEEoverridecommandlockouts

\usepackage[numbers, sort&compress, square]{natbib}
\usepackage{amsmath,amssymb,amsfonts}
\usepackage{algorithmic}
\usepackage{graphicx}
\usepackage{textcomp}
\usepackage{xcolor}
\def\BibTeX{{\rm B\kern-.05em{\sc i\kern-.025em b}\kern-.08em
    T\kern-.1667em\lower.7ex\hbox{E}\kern-.125emX}}

\usepackage[utf8]{inputenc} 
\usepackage[T1]{fontenc}    
\usepackage{hyperref}       
\usepackage{url}            
\usepackage{booktabs}       
\usepackage{amsfonts}       
\usepackage{nicefrac}       
\usepackage{microtype}      
\usepackage{enumitem}
\usepackage{amsmath}
\usepackage{subcaption}
\usepackage{float}
\newcommand{\ignore}[1]{\index{ignore}}
\usepackage{tabularx} 
\usepackage{booktabs}

\begin{document}

\title{Pixel-Based Similarities as an Alternative to Neural Data for Improving Convolutional Neural Network Adversarial Robustness}

\author{\IEEEauthorblockN{Elie Attias}
\IEEEauthorblockA{\textit{SEAS}\\
\textit{Harvard University}\\
Cambridge, MA, USA \\
elieattias@g.harvard.edu}
\and
\IEEEauthorblockN{Cengiz Pehlevan}
\IEEEauthorblockA{\textit{SEAS}\\
\textit{Harvard University}\\
Cambridge, MA, USA \\
cpehlevan@seas.harvard.edu}
\and
\IEEEauthorblockN{Dina Obeid}
\IEEEauthorblockA{\textit{SEAS}\\ 
\textit{Harvard University}\\
Cambridge, MA, USA  \\
dinaobeid@seas.harvard.edu}
}

\maketitle

\begin{abstract} \label{abstract}    
Convolutional Neural Networks (CNNs) excel in many visual tasks but remain susceptible to adversarial attacks—imperceptible perturbations that degrade performance. Prior research reveals that brain-inspired regularizers, derived from neural recordings, can bolster CNN robustness; however, reliance on specialized data limits practical adoption.
We revisit a regularizer proposed by Li et al.~(2019) that aligns CNN representations with neural representational similarity structures and introduce a data-driven variant. Instead of a neural recording–based similarity, our method computes a pixel-based similarity directly from images. This substitution retains the original biologically motivated loss formulation, preserving its robustness benefits while removing the need for neural measurements or task-specific augmentations.
Notably, this data-driven variant provides the same robustness improvements observed with neural data. Our approach is lightweight and integrates easily into standard pipelines. Although we do not surpass cutting-edge specialized defenses, we show that neural representational insights can be leveraged without direct recordings. This underscores the promise of robust yet simple methods rooted in brain-inspired principles, even without specialized data, and raises the possibility that further integrating these insights could push performance closer to human levels without resorting to complex, specialized pipelines.

\end{abstract}


\section{Introduction}
\label{introduction}
Convolutional Neural Networks (CNNs) have achieved high performance on a variety of visual tasks such as image classification and segmentation. Despite their remarkable success, these networks are notably brittle; even a small change in the input can significantly alter the network's output \cite{biggio2013evasion,szegedy2013}. Szegedy et al. \cite{szegedy2013} found that small perturbations, imperceptible to the human eye, can lead CNNs to misclassify images. These adversarial images pose a significant threat to computer-vision models.

Improving the robustness of CNNs against adversarial inputs is a major focus in machine learning. Various methods have been proposed, each with different levels of success and computational demands \cite{li2022review}. Some researchers have drawn inspiration from the mammalian brain, finding that deep neural networks trained to mimic brain-like representations are more resistant to adversarial attacks \citep{li2019learning, safarani2021towards, li2023robust}. Li et al. \cite{li2019learning} demonstrated that incorporating a regularizer into the loss function, which aligns the CNN's representational similarities \citep{kriegeskorte2008representational} with those of the mouse primary visual cortex (V1), significantly enhances the network's robustness to Gaussian noise and adversarial attacks. Using a loss term to steer models towards brain-like representations is referred to as neural regularization. However, a significant drawback of these methods is the reliance on neural recordings, which are often difficult to obtain and limit the methods' applicability.

In this work, we revisit the neural regularizer of \cite{li2019learning}, focusing on its use in CNNs for image classification—the task domain in which it was originally evaluated. We ask whether the underlying computational principles of this biologically inspired approach can be abstracted to improve network robustness without relying on large-scale neural recordings. The regularizer shapes CNN representations to match the representational similarity structure of a predictive model of mouse V1, which serves to denoise the inherently noisy neural measurements. 

We observe that the representational similarities produced by the predictive model in \cite{li2019learning} correlate strongly with simple pixel-level similarities. Motivated by this, we propose a simple and interpretable similarity-based regularizer that depends only on the regularization image dataset and does not require neural data. We evaluate the robustness of models trained with our method under black-box attacks, where the adversary has only query access, and find consistent improvements across many attacks; however, we also show that the method does not confer robustness against all black-box attack types, suggesting that capturing additional brain-inspired principles may be necessary. Our approach is flexible: it supports different regularization datasets—including the classification dataset itself—and works on both grayscale and color data. We further assess robustness to common corruptions using CIFAR-10-C \cite{hendrycks2019benchmarking}. By analyzing the Fourier spectra of minimal adversarial perturbations obtained via decision-based Boundary Attacks \cite{brendel2017decision}, we find that our regularizer primarily protects against high-frequency perturbations. Finally, we show that our method is computationally light, requiring only a small regularization batch and a modest number of regularizing images, and provide a simple heuristic for selecting the regularization hyperparameters.

Our work demonstrates that a brain-inspired regularizer can enhance model robustness without large-scale neural recordings. This contributes to the broader use of biologically-inspired loss functions to improve artificial neural networks' performance. The resulting method is a simple, computationally inexpensive regularizer that performs well across a wide range of scenarios.

\section{Related work}\label{related_works}

\noindent\textbf{Adversarial attacks.} Identifying adversarial examples that can mislead modern neural networks is an active area of research, with a growing number of attack methods introduced in recent years \cite{szegedy2013, hinton2015distilling, moosavi2016deepfool, brendel2017decision, madry2017towards, guo2019simple}.
In this work, we focus on black-box attacks, which do not require access to model gradients or parameters and therefore better reflect real-world threat models. We evaluate robustness against four classes of attacks: random noise, transfer-based attacks, decision-based attacks, and common corruptions.

Random noise attacks apply perturbations sampled from distributions such as Gaussian, Uniform, or Salt-and-Pepper noise. Transfer-based attacks craft adversarial examples on a substitute model—here, an unregularized baseline—and apply them to the target model, leveraging the transferability of adversarial perturbations across architectures \cite{papernot2016transferability}. We generate these perturbations using the Fast Gradient Sign Method (FGSM) \cite{goodfellow2014explaining}. Decision-based attacks rely solely on model outputs. We evaluate robustness to the Boundary Attack \cite{brendel2017decision}, which iteratively reduces an initial large perturbation while keeping the example adversarial, and to SimBA \cite{guo2019simple}, a query-efficient black-box attack that applies coordinate-wise perturbations in pixel or DCT space and accepts updates that reduce the model’s confidence in the correct class. Finally, common corruptions correspond to real-world distortions (e.g., motion blur) and are taken from the CIFAR-10-C benchmark \cite{hendrycks2019benchmarking}. 
Additional implementation details for the attack procedures are provided in the supplementary material.\\

\vspace{-2mm}
\noindent\textbf{Adversarial training.} As attacks have advanced, numerous defenses have been proposed \cite{goodfellow2014explaining, bhagoji2018enhancing, diffenderfer2021winning, kireev2022effectiveness}.
The most widely used among these is adversarial training, which augments training batches with adversarial examples \cite{goodfellow2014explaining, madry2017towards}. The Projected Gradient Descent (PGD) attack \cite{madry2017towards}, a multi-step extension of FGSM, is commonly employed due to its effectiveness in generating strong adversarial samples that improve model robustness.\\

\vspace{-2mm}
\noindent\textbf{Neural regularization.} Recent work has examined using neural data to regularize deep networks by encouraging brain-like internal representations \cite{li2019learning, safarani2021towards}. This can be achieved by adding a penalty term that aligns either representational similarities or neural activations with those observed in early visual cortex. For example, \cite{li2019learning} encouraged CNN representations to match representational similarity patterns from mouse primary visual cortex (V1) \cite{kriegeskorte2008representational}, while \cite{safarani2021towards} trained networks to predict macaque V1 responses to natural stimuli. A key limitation of these approaches is their reliance on large-scale neural recordings, which are difficult and expensive to obtain.

\section{Motivation and Method}\label{method}

To increase the robustness of artificial neural networks to adversarial attacks, one research direction focuses on extracting and applying computational concepts from the mammalian brain.
In \cite{li2019learning}, adding a neural regularizer term to the training loss was shown to improve the adversarial robustness of CNNs on image classification tasks. The full loss is
\begin{equation}
\label{total_loss}
L = L_{\text{task}} + \alpha L_{\text{sim}},
\end{equation}
where $\alpha$ controls the regularization strength. For a pair of images $(i,j)$, the similarity loss is
\begin{equation}
\label{sim_loss}
L_{\text{sim}} = \sum_{i\neq j} \left (\operatorname{arctanh}(S_{ij}^{\text{CNN}}) -\operatorname{arctanh}(S_{ij}^{\text{neural}}) \right )^2,
\end{equation}
where $S^{\text{neural}}_{ij}$ denotes the pairwise cosine similarity between neural representations of images $i$ and $j$, and $S^{\text{CNN}}_{ij}$ is the corresponding similarity between CNN features. Following \cite{li2019learning}, we compute $S^{\text{CNN}}_{ij}$ by combining similarities across $K$ equally spaced convolutional layers using trainable, non-negative weights $\gamma_l$ obtained from a softmax (so $\sum_l\gamma_l = 1$):
\begin{equation}
S^{\text{CNN}}_{ij} = \sum_{l} \gamma_l \, S^{\text{CNN-}l}_{ij},
\end{equation}
where $S^{\text{CNN-}l}_{ij}$ is the mean-subtracted cosine similarity at layer $l$. The trainable weights $\gamma_l$ allow the model to select which layers should be aligned more strongly with neural representations.
In \cite{li2019learning}, the authors trained a ResNet \cite{he2016deep} on grayscale CIFAR-10 and CIFAR-100 while regularizing to match representational similarities derived from mouse V1 responses to grayscale ImageNet images. Because neural recordings are noisy, these similarities were not computed directly from the measured neural activity. Instead, they were obtained from a predictive model trained to estimate V1 responses \cite{sinz2018stimulus, walke2018inception}. We denote these model-derived target similarities as $S^{\text{neural-pred}}$. The predictive model is a three-layer CNN with skip connections and incorporates behavioral covariates such as pupil position, pupil size, and running speed. Additional details are provided in the supplementary material.

Training alternates between batches from the classification dataset to compute $L_{\text{task}}$ and batches of image pairs from the regularization dataset to compute $L_{\text{sim}}$, after which the full loss $L$ is backpropagated.\\

\begin{figure}[t]
    \centering
        \includegraphics[width=0.95\linewidth]{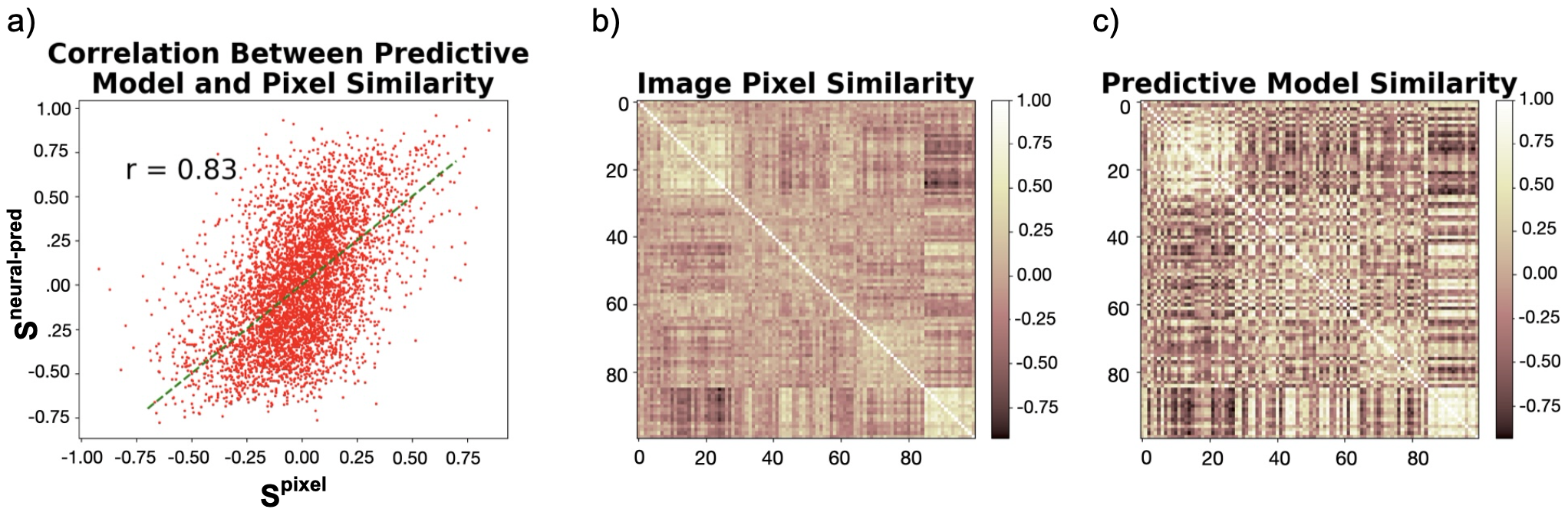}
        \caption{Relationship between image pixel similarity and V1-based representational similarity. (a) Correlation between pixel similarity ($S^{pixel}$) and V1-based representational similarity ($S^{\text{neural-pred}}$) computed from the predictive model of \cite{li2019learning} (Pearson’s $r = 0.83$). Pixel similarity is measured as the cosine similarity between flattened, mean-subtracted, and normalized pixel vectors. (b) Pixel similarity matrix. (c) V1-based representational similarity matrix, obtained by averaging similarities across predictive models trained on six distinct neural scans. Both matrices share similar global structure, though the V1-based matrix spans a wider dynamic range. Additional details are provided in the supplementary material.}
        \label{fig:pixel-neural-correlation}
\end{figure}

\noindent \textbf{Pixel-based insight.}
The primary visual cortex (V1) is the first cortical area for visual processing, where neurons respond to oriented bars and other low-level features, as opposed to the more complex features encoded in higher visual areas. Comparing image pixel similarity to the V1-based similarity computed using the predictive model in \cite{li2019learning} (Fig.\,\ref{fig:pixel-neural-correlation},\,a) shows a high correlation (Pearson’s $r=0.83$). Qualitatively, the image pixel similarity matrix and the V1-based similarity matrix (Fig.\,\ref{fig:pixel-neural-correlation},\,b,\,c) exhibit the same global structure (e.g., image pairs deemed similar by one measure are also similar by the other), although the V1-based matrix spans a wider dynamic range. This suggests that early cortical processing roughly preserves an underlying pixel-similarity structure.

Having observed this relationship, we next evaluate the use of image pixel similarity $S^{pixel}$—measured as the cosine similarity between flattened, mean-subtracted, and normalized pixel vectors—in the regularizer $L_{\text{sim}}$ in place of $S^{\text{neural}}$, following the experimental setup of \cite{li2019learning}. As shown in Fig.\,\ref{comparing-li-et al}, using $S^{pixel}$ yields some improvement in robustness to Gaussian noise and the decision-based Boundary Attack; however, this improvement does not extend to other attacks such as transfer-based FGSM. Additional results are provided in the supplementary material.\\
\newline \noindent \textbf{Thresholded pixel-based similarity}.
Motivated by this observation, and keeping the original loss formulation intact, we replace $S_{ij}^{\text{neural}}$ with a thresholded pixel-based similarity $S_{ij}^{Th}$, defined as
\begin{equation}
S_{ij}^{Th} =
\begin{cases}
     1, & \text{if } S_{ij}^{\text{pixel}} > Th,\\
    -1, & \text{if } S_{ij}^{\text{pixel}} < -Th,\\
     0, & \text{if } |S_{ij}^{\text{pixel}}| \le Th,
\end{cases}
\label{S_boosted}
\end{equation}
where $Th\in(0,1)$ is a tunable hyperparameter and $S_{ij}^{\text{pixel}}$ is the cosine similarity between mean-subtracted, normalized pixel vectors. In practice, the values $1$ and $-1$ in \eqref{S_boosted} are replaced by $1-\epsilon$ and $-1+\epsilon$, respectively, to avoid divergence of the $\operatorname{arctanh}$ function in \eqref{sim_loss}. Although we use the same loss $L_{\text{sim}}$ as in \eqref{sim_loss}, we note that the $\operatorname{arctanh}$ operation is not essential in this thresholded setting. Enforcing these thresholded targets $S_{ij}^{Th}$ pushes the CNN to produce orthogonal representations for image pairs whose pixel similarity lies in $[-Th,Th]$, and aligned representations otherwise. Empirically, the earliest convolutional layer absorbs most of the effect, treating highly similar images ($|S_{ij}^{\text{pixel}}|>Th$) as small perturbations of one another. In Section~\ref{alpha,th}, we describe a simple method for selecting the hyperparameters $Th$ and $\alpha$.
In the supplementary material, we show the contribution of each term in \eqref{S_boosted} to robustness.

Finally, the proposed regularization method offers several advantages: (1) it does not require large-scale neural recordings, (2) it relies solely on the original datasets and requires no data distortions or augmentations, and (3) it is computationally inexpensive (see Section~\ref{comp-advantage}).

\section{Datasets and Experiments}\label{results}

We train ResNets \cite{he2016deep} on image classification tasks using the loss
$L = L_{\text{task}} + \alpha\, L_{\text{sim}}$ \eqref{total_loss}, where the regularization term
$L_{\text{sim}}$ \eqref{sim_loss} is computed using the thresholded pixel-based similarity
$S_{ij}^{Th}$ in place of $S_{ij}^{\text{neural}}$. In our experiments, we primarily use
$(\alpha, Th) = (10, 0.8)$. Additional information on the values of the hyperparameters for each classification–regularization dataset combination, as well as the learned layer weights
$\gamma_l$ (defined in Section~\ref{method}), is provided in the supplementary material.

After training, we evaluate robustness against several black-box attacks, including Gaussian noise, transfer-based FGSM \cite{goodfellow2014explaining}, and the Boundary Attack \cite{brendel2017decision}. We additionally evaluate robustness to the SimBA attack \cite{guo2019simple}, as well as to common corruptions using the grayscale CIFAR-10-C dataset. See Section \ref{related_works} for more details.

Following the experimental setup of \cite{li2019learning}, we train a ResNet-18 to classify
grayscale CIFAR-10 while regularizing with grayscale ImageNet, and then extend the experiments
to color CIFAR-10 with regularization datasets drawn from color CIFAR-10, CIFAR-100, and
ImageNet. Similar trends are observed when training ResNet-34 on CIFAR-100, as well as when
using MNIST and Fashion-MNIST as both classification and regularization datasets. These results,
together with full experimental details and implementation specifics, are included in the
supplementary material. All codes are available at \href{https://github.com/elieattias1/pixel-reg}{https://github.com/elieattias1/pixel-reg}.

\subsection{Evaluating robustness}\label{robustness_results}
\begin{figure}[t]
\centering
\includegraphics[width=0.99\linewidth]{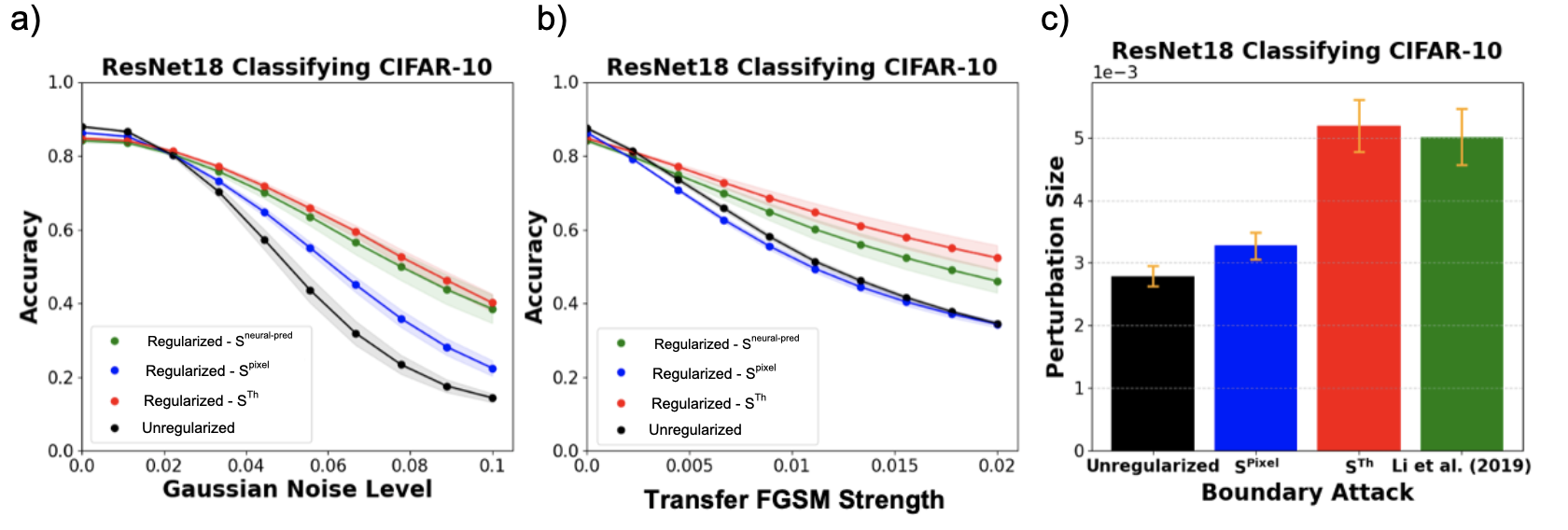}
\caption{Robustness of ResNet18 trained to classify grayscale CIFAR-10 and regularized with images from grayscale ImageNet dataset. Robustness to (a) Gaussian noise, (b) transfer-based FGSM \cite{goodfellow2014explaining}, and (c) decision-based Boundary Attack \cite{brendel2017decision}. For comparison, results from using different regularization targets in $L_{sim}$ are shown : $S^{pixel}$, $S^{Th}$ and $S^{neural-pred}$-the neural-based targets as in \cite{li2019learning} (see Section \ref{method}). For the decision-based  Boundary Attack, we compute the median squared $L_2$ perturbation size per pixel, averaged across 1000 images, and 5 repeats. Error shades represent the SEM across seven seeds per model.}
\label{comparing-li-et al}
\end{figure}

We evaluate the robustness of models regularized using $S^{Th}$ on grayscale CIFAR-10, with grayscale ImageNet used as the regularization dataset, following the setup in \cite{li2019learning}. Under Gaussian noise perturbations, the regularized models show a substantial improvement in robustness compared to unregularized models (Fig.\,\ref{comparing-li-et al},\,a), and perform similarly to the neural-regularized models in \cite{li2019learning}. Results for robustness to Uniform and Salt-and-Pepper noise are provided in the supplementary material.
We further evaluate robustness to stronger black-box attacks, namely transfer-based FGSM \cite{goodfellow2014explaining} and the decision-based Boundary Attack \cite{brendel2017decision}. As shown in Fig.\,\ref{comparing-li-et al},\,b,c, models regularized with $S^{Th}$ achieve increased robustness to both attacks. For the Boundary Attack, larger final perturbation magnitudes indicate greater robustness; here too, $S^{Th}$-regularized models match the performance of neural regularized models \cite{li2019learning}.

In all panels of Fig.\,\ref{comparing-li-et al}, we also include results 
obtained by regularizing directly with image pixel similarity $S^{pixel}$ to allow direct comparison. Overall, these experiments demonstrate that regularizing with $S^{Th}$ yields robustness comparable to neural regularization \cite{li2019learning} while requiring no neural data and relying solely on the original, unaugmented regularization dataset.

\begin{figure}[bt]
    \centering
    \includegraphics[width=0.99\linewidth]{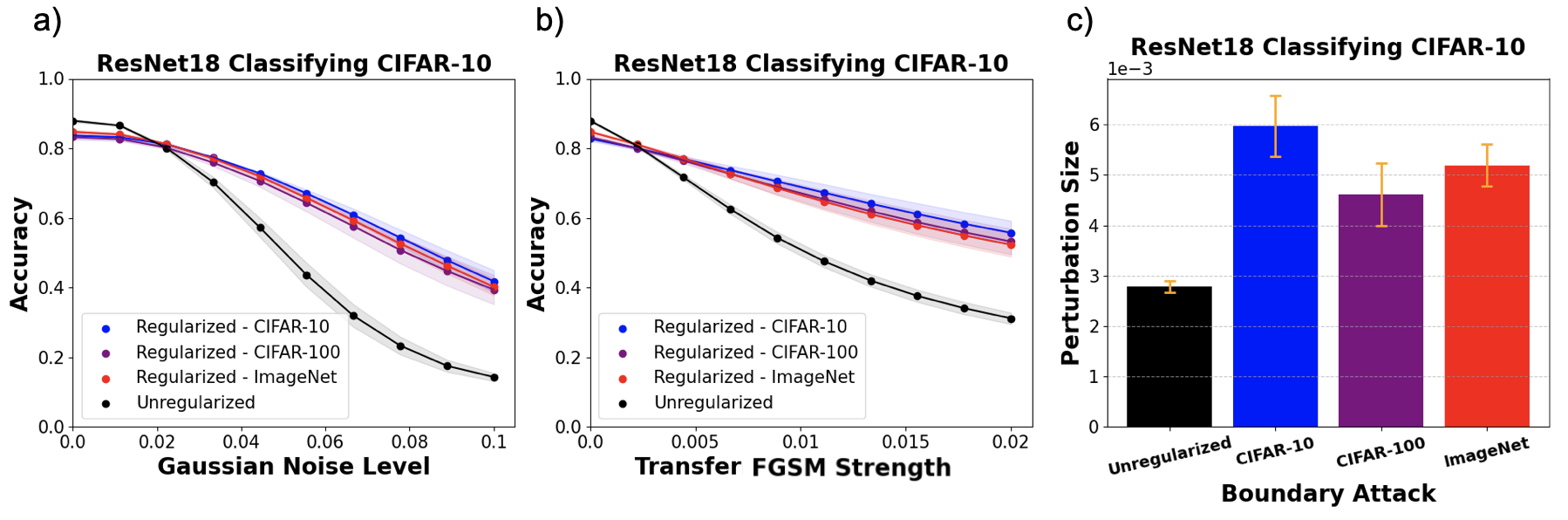}
    \caption{Robustness of a ResNet18 trained to classify grayscale CIFAR-10 regularized on grayscale images from different datasets : CIFAR-10 (blue), CIFAR-100 (purple) or ImageNet (red). For the decision-based Boundary Attack, we compute the median $L_2$ perturbation size, averaged across 1000 images, and 5 repeats. Error shades/bars represent the SEM across seven seeds per model. The same ($\alpha$,$Th$) values are used in training all models i.e., for all regularization datasets (see the supplementary material).}  
    \label{fig:dataset-combinations}  
\end{figure}

\begin{figure}[t]
\centering
\includegraphics[width=0.99\linewidth]{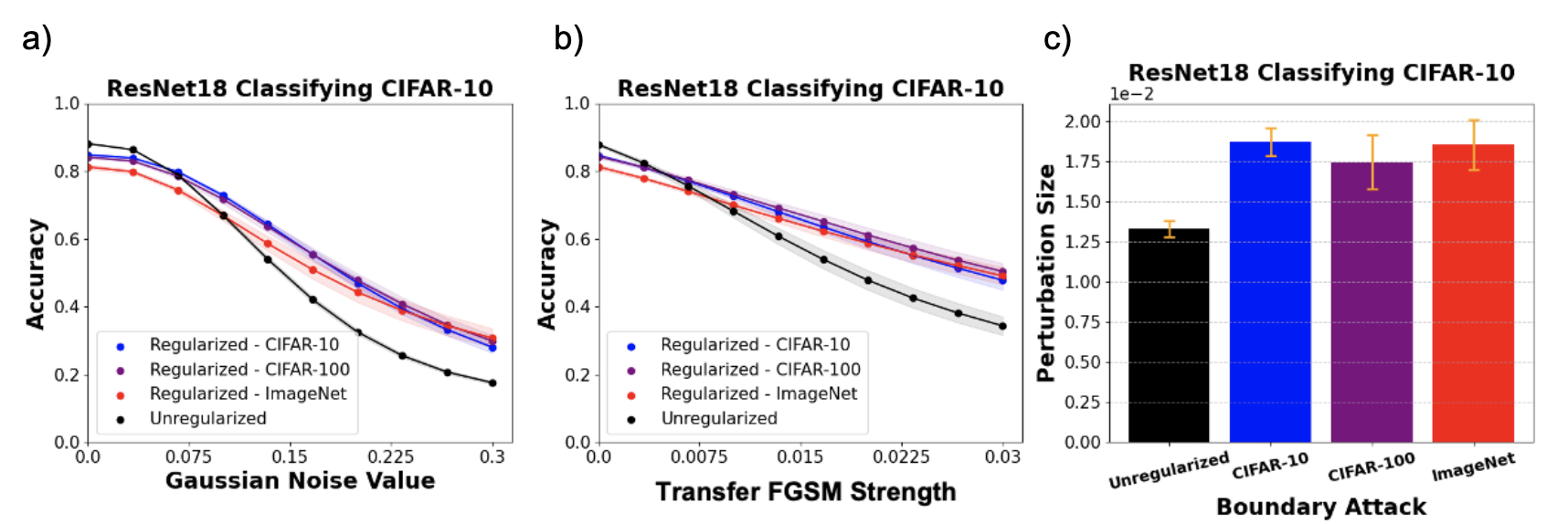}
\caption{Robustness of a ResNet18 trained to classify color CIFAR-10 regularized on color images from different datasets : CIFAR-10 (blue), CIFAR-100 (purple) or ImageNet (red). For the decision-based Boundary Attack, we compute the median $L_2$ perturbation size, averaged across 1000 images, and 5 repeats. Error shades/bars represent the SEM across seven seeds per model. }
\label{fig:color-cifar10}
\end{figure}

\begin{figure}
    \centering
    \includegraphics[width=0.9\linewidth]{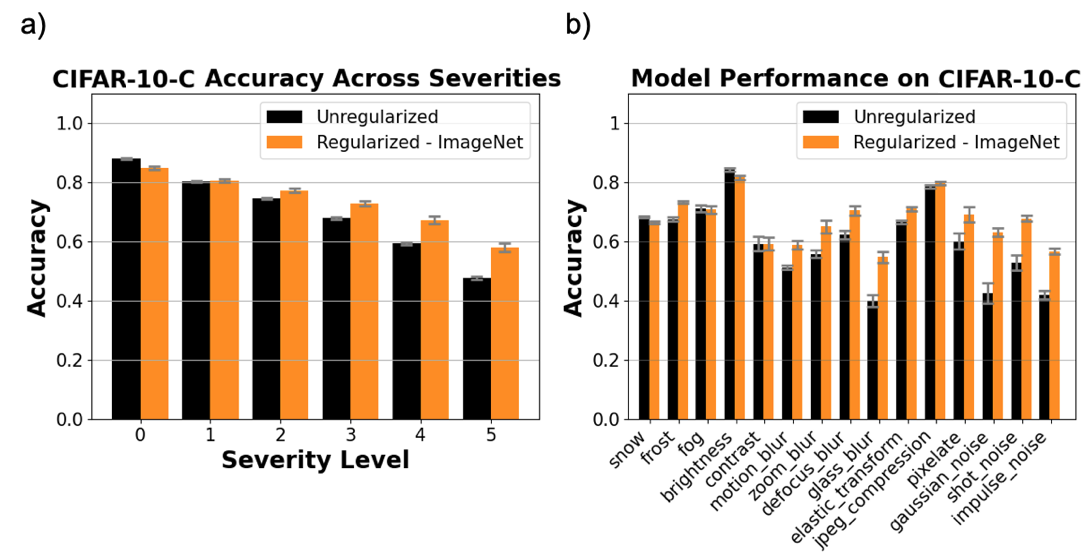}
    \caption{Robustness to grayscale CIFAR-10-C corruptions \cite{hendrycks2019benchmarking}. Results for regularized and unregularized models. (a) Accuracy across severity levels, averaged over all 15 corruption types. (b) Performance on the 15 individual corruptions at severity~4. Error bars show SEM over seven seeds. Models are ResNet-18 trained on grayscale CIFAR-10 regularized with grayscale ImageNet.} 
    \label{fig:C10_IM-CC}  
\end{figure}

\begin{figure}
    \centering
    \includegraphics[width=0.4\textwidth]{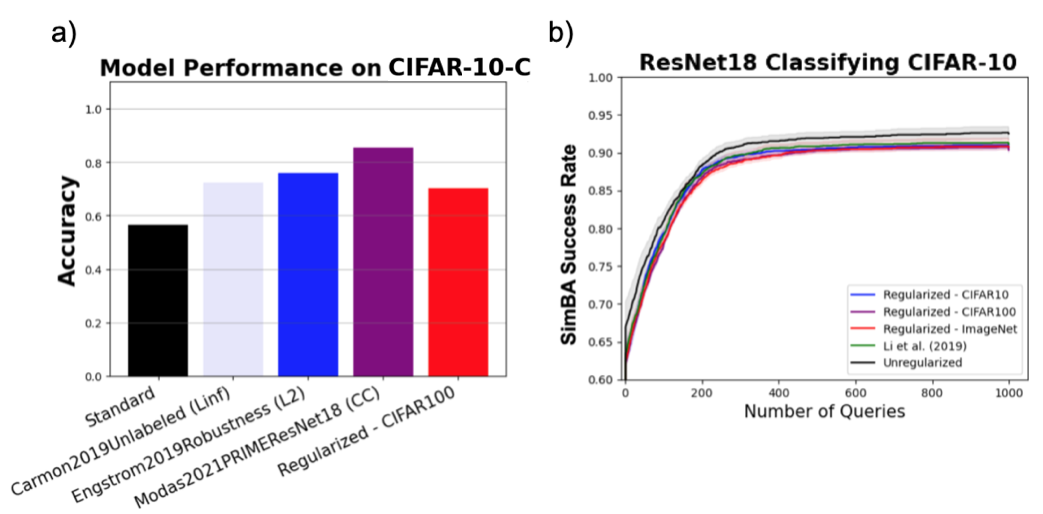}
    \caption{\small (a) Benchmarking our method against other adversarially trained networks. (b) Robustness against SimBA \cite{guo2019simple}: ResNet18 trained to classify grayscale CIFAR-10 regularized with grayscale CIFAR-10, CIFAR-100, or ImageNet. We also show results from models regularized using neural-data \cite{li2019learning}.}
    \label{fig:benchmark-simba}
\end{figure}

Our method is flexible with respect to the choice of regularization dataset. 
Although we did not evaluate an exhaustive range of publicly available dataset combinations, the ones we tested all show robustness improvements, although some qualitative differences can be observed across dataset combinations (Figs.~\ref{fig:dataset-combinations} and \ref{fig:color-cifar10}). 
Figure~\ref{fig:dataset-combinations} shows the performance of a ResNet-18 trained to classify grayscale CIFAR-10 and regularized using grayscale images from CIFAR-10, CIFAR-100, or ImageNet, evaluated under three black-box attacks (Gaussian noise, transfer-based FGSM, and the decision-based Boundary Attack). 
All regularization datasets yield robustness gains relative to the unregularized model. 
The supplementary material includes results for other classification–regularization dataset combinations.
%
Our previous results were obtained using grayscale datasets, chosen to enable direct comparison with the results from using the neural regularizer in \cite{li2019learning}. Here, we demonstrate that our method also extends naturally to color datasets, which are more commonly used in practical computer vision settings. 
Figure~\ref{fig:color-cifar10} shows results for a ResNet-18 trained to classify color CIFAR-10 and regularized using color images from CIFAR-10, CIFAR-100, or ImageNet. All regularization datasets yield an increase in robustness relative to the unregularized model. The supplementary material includes results for ResNet-34 trained to classify color CIFAR-100.
%
Regularized models are also more robust than their unregularized counterparts on common corruptions. 
We evaluate robustness on the grayscale CIFAR-10-C benchmark \cite{hendrycks2019benchmarking}, which contains grayscale CIFAR-10 images corrupted with distortions commonly encountered in real-world vision systems. Evaluating across multiple corruption types and severity levels is essential, as they simulate practical deployment conditions. Figure~\ref{fig:C10_IM-CC} shows the performance of a ResNet-18 trained to classify grayscale CIFAR-10 and regularized using grayscale ImageNet. 
Figure~\ref{fig:C10_IM-CC}\,a reports accuracy averaged across all 15 corruption types at each severity level. Figure~\ref{fig:C10_IM-CC}\,b shows performance on each individual corruption at severity~4. In all cases, regularization improves robustness relative to the unregularized model.
We also compare our method—using a ResNet-18 trained on color CIFAR-10 and regularized with color CIFAR-100—against three widely used adversarially trained models from the RobustBench leaderboard \cite{croce2020robustbench}: Carmon 2019 ($L_{\infty}$) \cite{carmon2019unlabeled}, Engstrom 2019 ($L_2$) \cite{robustness}, and Modas 2021 PRIME-ResNet-18 \cite{modas2022prime}. 
Performance is evaluated on CIFAR-10-C at severity~5, and we also include the “standard” RobustBench baseline for reference (Fig.~\ref{fig:benchmark-simba}\,a). 
As expected, our method does not match state-of-the-art adversarially trained models, which are specifically optimized for adversarial robustness.

Finally, we note that our method does not improve robustness against all black-box attacks. 
Figure~\ref{fig:benchmark-simba}\,b shows that it is not effective against SimBA \cite{guo2019simple}; however, its performance closely matches that of the neural-regularized model from \cite{li2019learning}. 
The aim of our method is not to outperform adversarially trained networks, but rather to demonstrate that a neural-regularizer–inspired approach can achieve comparable behavior to neural regularization without requiring expensive neural recordings.

\subsection{Frequency decomposition of adversarial perturbations and common corruptions}
\begin{figure}
    \centering  \includegraphics[width=1\linewidth]{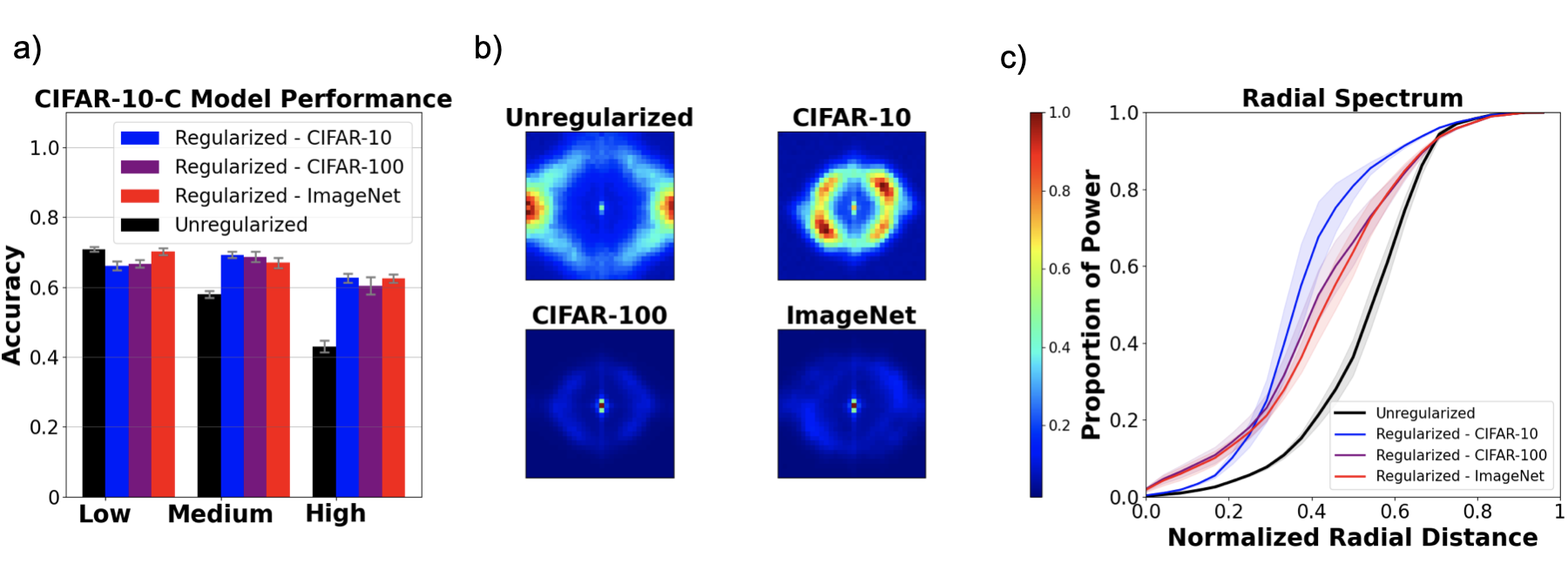}
    \caption{Frequency perspective on robustness. Results of ResNet18 trained to classify grayscale CIFAR-10 regularized on grayscale images from : CIFAR-10, CIFAR-100 or ImageNet. (a) Robustness of regularized models evaluated on grayscale CIFAR-10-C at severity 4, categorized by the frequency range of each corruption. (b) Fourier power spectrum for the mean minimal corruption required to flip a model's decision. (c) Radial Spectrum of minimal perturbation required to mislead models, as provided by a decision-based Boundary Attack \cite{brendel2017decision} - using 10k steps.  The error bars (left panel) and shaded areas (right panel) represent the SEM across seven and four seeds per model respectively.}
    \label{fig:cifar10-frequencies}   
\end{figure}

To understand the strengths and weaknesses of our regularization method, we investigate the frequency components present in the minimal perturbation required to flip the decision of unregularized and regularized models which we compute \textit{via} a decision-based Boundary Attack \cite{brendel2017decision}. We observe that models regularized using $S^{Th}$ rely more on low frequency information than their unregularized counterparts (Fig.~\ref{fig:cifar10-frequencies}\,b,\,c). We further evaluate our regularized model performance on grayscale CIFAR-10-C \cite{hendrycks2019benchmarking} following the approach described in \cite{li2023robust}, where we categorize the 15 corruptions in CIFAR-10-C into Low, Medium, and High frequency based on their spectra (see Fig.~\ref{fig:cifar10-frequencies}\,a and the supplementary material). Results are shown for ResNet18 trained to classify  grayscale CIFAR-10 and regularized using grayscale images from CIFAR-10, CIFAR-100 or ImageNet datasets. Our results show that regularized models outperform unregularized ones, especially on high-frequency corruptions, confirming our findings. Such a reliance on low-frequency information has also been observed in models subjected to neural regularization as explained in \cite{li2023robust}.
\section{Computational advantage}\label{comp-advantage}
\begin{figure}
    \centering
    \includegraphics[width=1\linewidth]{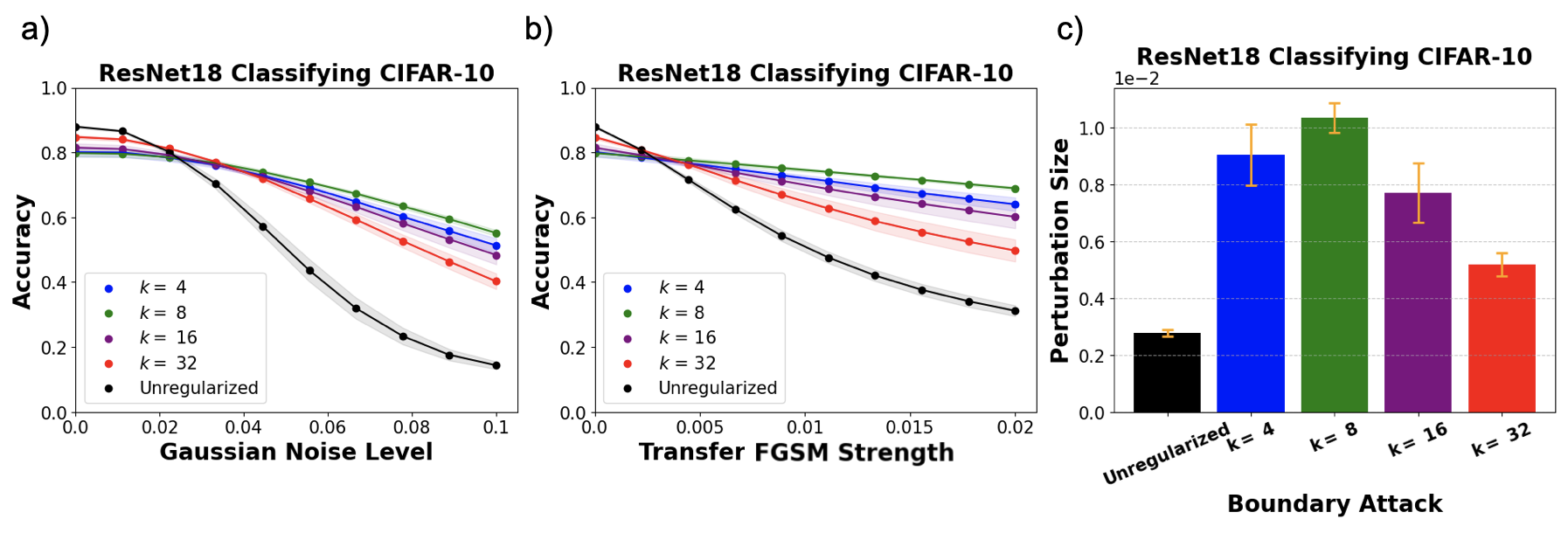}
    \caption{Robustness of a ResNet18 trained to classify grayscale CIFAR-10 regularized on grayscale images from ImageNet dataset is shown  for different regularization batch sizes, $k \in \{4, 8, 16, 32\}$. For the decision-based Boundary Attack, we compute the median $L_2$ perturbation size, averaged across 1000 images, and 5 repeats. Error shades/bars represent the SEM across seven seeds per model. }  
    \label{fig:regularization-batch_size}  
\end{figure}

Beyond its simplicity, our regularization method is computationally inexpensive. 
A regularization batch containing $k$ image pairs adds only $2k$ additional forward passes per training step. 
As shown in Fig.~\ref{fig:regularization-batch_size}, the method remains effective across a wide range of batch sizes (4, 8, 16, 32). Thus, smaller batch sizes can be used to further reduce the extra training time.

Second, although the number of target similarities scales as $\binom{N}{2}$ for $N$ regularization images, we find that only a small number of images are sufficient. While our main experiments use $N = 5000$ images (yielding roughly $12\times 10^6$ pairs; see the supplementary material), Fig.~\ref{fig:regularization-reg_image_num} shows that even using $N \in \{100, 1000\}$ already provides meaningful robustness gains against black-box attacks. 

Finally, our method relies solely on the original image datasets and does not require additional distortions or data augmentations during training.
\begin{figure}
    \centering
    \includegraphics[width=0.94\linewidth]{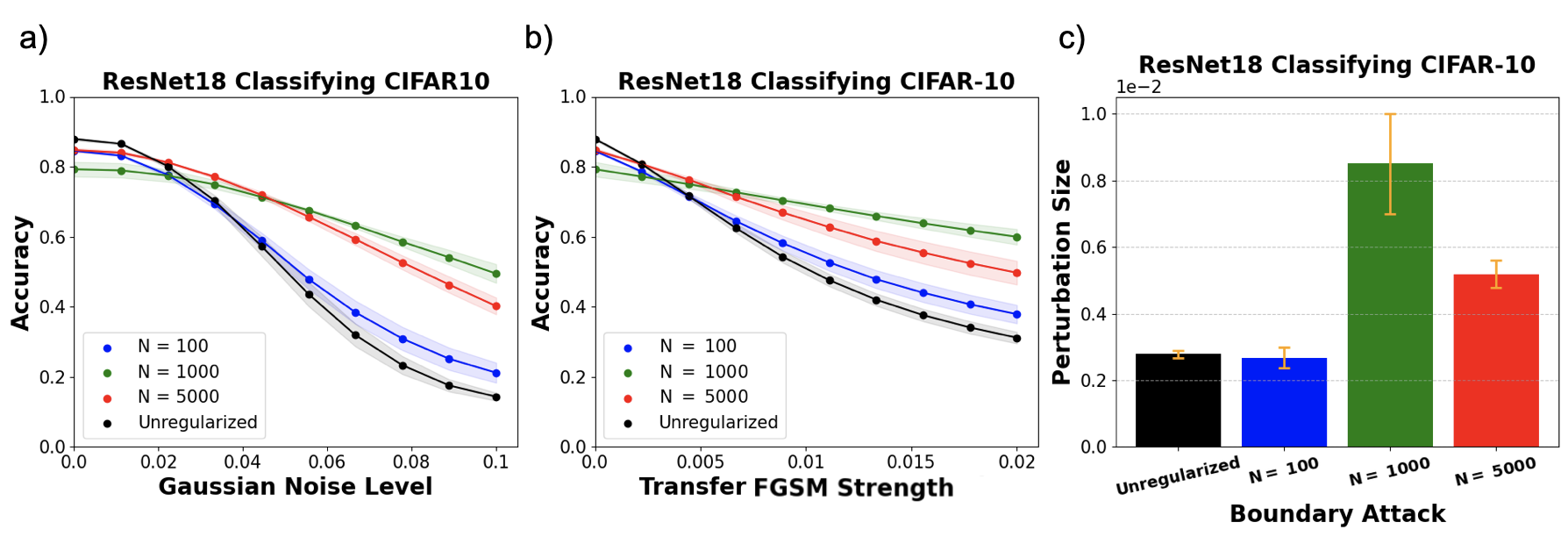}
    \caption{Robustness of a ResNet18 trained to classify grayscale CIFAR-10 regularized on grayscale images from ImageNet dataset is shown for different number of regularization images. For the decision-based Boundary Attack, we compute the median $L_2$ perturbation size, averaged across 1000 images, and 5 repeats. Error shades/bars represent the SEM across seven seeds per model. }  
    \label{fig:regularization-reg_image_num}  
\end{figure}

\section{Hyperparameter selection}\label{alpha,th}
\begin{figure}
    \centering
\includegraphics[width=0.6\linewidth]{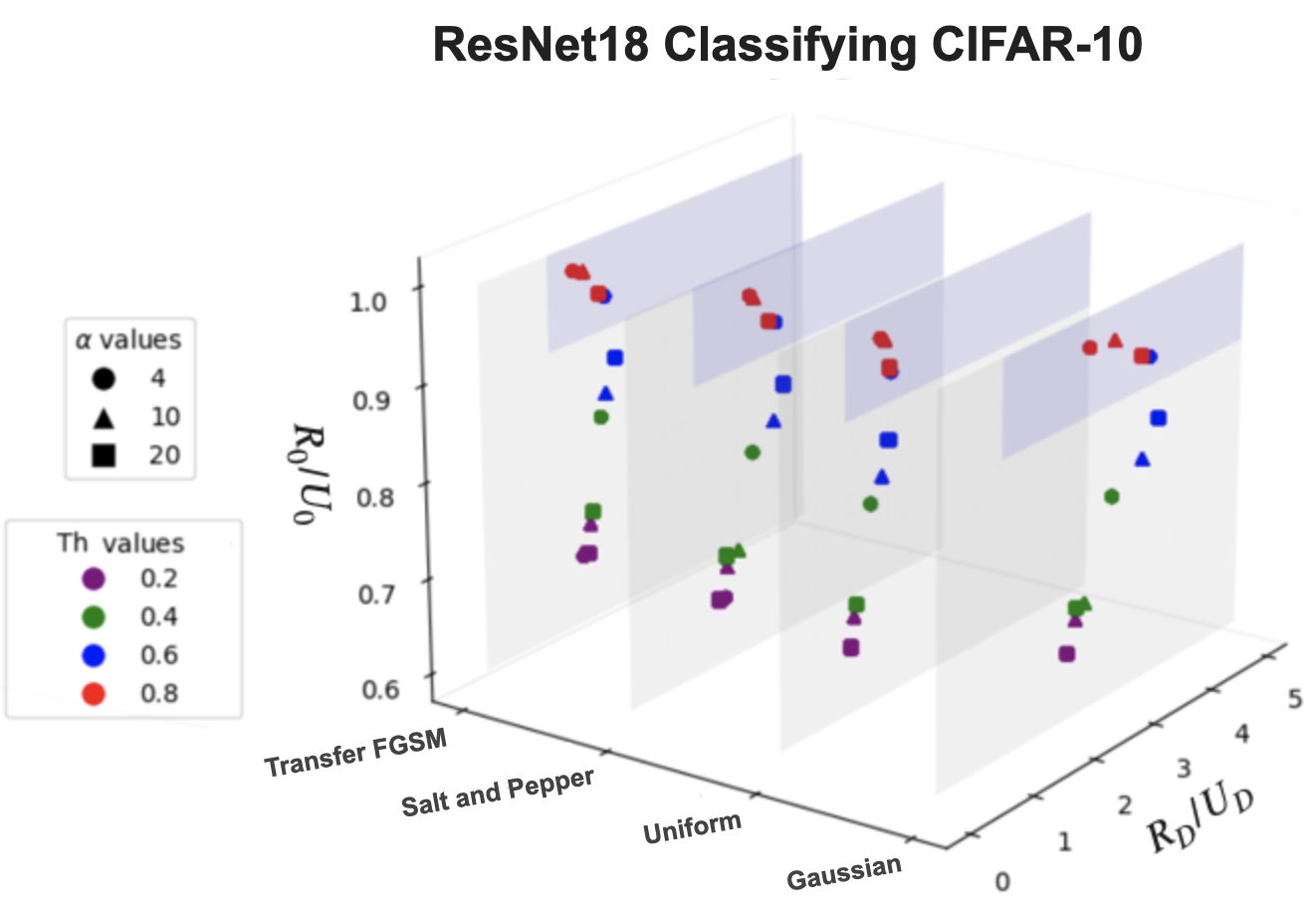}
    \caption{Evaluating ($\alpha$,$Th$) pairs under multiple attacks. ResNet-18 trained on grayscale CIFAR-10 regularized with grayscale ImageNet. Each plane shows results for a different black-box attack; blue regions mark “acceptable” models ($R_D/U_D>1$ and $R_0/U_0\ge0.9$--user selected criterion). We use $\epsilon = 0.1$ for random-noise attacks and $\epsilon = 0.02$ for transfer-based FGSM. Results are averaged over seven seeds.}
    \label{fig:trade-off}  
\end{figure}

An important question is how to select the hyperparameters $\alpha$ and $ Th$. We propose a criterion to select those hyperparameters, as follows. A suitable pair should (1) be such that the resulting model has an 'acceptable' accuracy on the distortion-free dataset, and (2) showcases an increase in robustness to adversarial attacks. To properly define what we mean by this, we define $R$ and $U$ as the regularized vs. unregularized accuracies on clean ($O$) or distorted ($D$) inputs. The ratios $\frac{R_0}{U_0}$ and $\frac{R_D}{U_D}$ reflect how our regularization affects the model's accuracy at zero and high distortion levels. To meet condition (1), we require that $\frac{R_0}{U_0} \geq A_0$, where $A_0$ is user-defined. We select $A_0 = 0.9$. Condition (2) is simply met by requiring that $\frac{R_D}{U_D} > 1$. 
We can visualize the performance of a model by plotting $\frac{R_0}{U_0}$ \textit{vs} $\frac{R_D}{U_D}$ for each ($\alpha$, $Th$) pair. This allows the user to select the hyperparameter pair based on the selection criterion that they choose. In Fig.\,\ref{fig:trade-off} we show such a plot for different adversarial attacks (the gray shaded planes); the blue shaded region in each plane represents the region where conditions (1) and (2) are met. 

In our (limited) experiments, we also observed that the trends across different attacks tend to align (Fig.~\ref{fig:trade-off}). Consequently, a quick Gaussian-noise evaluation is often sufficient to screen viable $(\alpha, Th)$ pairs before conducting more expensive robustness tests.

\section{Conclusion and Discussion}\label{conclusion}
Extracting computational principles from the brain to advance AI is a central long-term goal in neuroscience. 
Motivated by this objective, we revisited the brain-inspired regularization method of \cite{li2019learning}, which aligns CNN representations with neural representations via a similarity-based penalty derived from large-scale neural recordings. 
We identified the core mechanism underlying this regularizer and proposed a simple dataset-based variant that achieves comparable robustness without requiring neural data. Our formulation also provides an intuitive interpretation of the regularizer.

We demonstrated that our method improves robustness to a range of black-box attacks (Sections~\ref{method} and~\ref{robustness_results}), and that it scales to both grayscale and color datasets, including CIFAR-10 and CIFAR-100 (supplementary material). 
It supports multiple combinations of classification and regularization datasets and increases robustness to common corruptions, as shown using grayscale CIFAR-10-C. 
A Fourier analysis of minimal perturbations obtained via the decision-based Boundary Attack showed that perturbations for the regularized model contain relatively more low-frequency components than those of the unregularized model. 
Consistently, our method performs better against corruptions dominated by high-frequency distortions, in agreement with findings reported for neural-regularized models in \cite{li2023robust}. 
We also showed that the method is computationally light (Section~\ref{comp-advantage}) and introduced a simple procedure for selecting the hyperparameters $\alpha$ and $Th$ (Section~\ref{alpha,th}). 
Further analyses of the components of $S^{Th}$ are provided in the supplementary material.

Despite being inspired by a biologically motivated loss that originally required large-scale neural recordings, we demonstrated that similar robustness benefits can be obtained without any neural data. 
Our method is simple, requires no data augmentations, is flexible in the choice of regularization dataset, and performs well with small regularization batch sizes and comparatively few images used to construct $S^{Th}$ \eqref{S_boosted}. 
This work takes a step towards understanding the main working principles behind neural regularizers and toward developing practical, accessible robustness methods for the broader machine-learning community.

A limitation of our approach is its reduced effectiveness against some black-box attacks (Fig.~\ref{fig:benchmark-simba}\,b) and specific common corruptions (Fig.~\ref{fig:C10_IM-CC}\,b; Fig.~\ref{fig:cifar10-frequencies}\,a), suggesting that additional brain-inspired principles may be needed. 
While our method does not match the robustness of state-of-the-art adversarial defenses \cite{croce2020robustbench}, our aim is not to compete with such methods, but to show that the key insights from the neural regularizer in \cite{li2019learning} can be preserved without the need for expensive neural recordings.

\section*{Acknowledgments}
C.P. is supported by an NSF CAREER Award (IIS-2239780), DARPA grants DIAL-FP-038 and AIQ-HR00112520041, the Simons Collaboration on the Physics of Learning and Neural Computation, and the William F. Milton Fund from Harvard University. This work has been made possible in part by a gift from the Chan Zuckerberg Initiative Foundation to establish the Kempner Institute for the Study of Natural and Artificial Intelligence.

\begin{small}
\bibliographystyle{ieeetr}
\bibliography{adversarial}

@article{li2019learning,
  title={Learning from brains how to regularize machines},
  author={Li, Zhe and Brendel, Wieland and Walker, Edgar and Cobos, Erick and Muhammad, Taliah and Reimer, Jacob and Bethge, Matthias and Sinz, Fabian and Pitkow, Zachary and Tolias, Andreas},
  journal={Advances in neural information processing systems},
  volume={32},
  year={2019}
}

@inproceedings{biggio2013evasion,
  title={Evasion attacks against machine learning at test time},
  author={Biggio, Battista and Corona, Igino and Maiorca, Davide and Nelson, Blaine and {\v{S}}rndi{\'c}, Nedim and Laskov, Pavel and Giacinto, Giorgio and Roli, Fabio},
  booktitle={Machine Learning and Knowledge Discovery in Databases: European Conference, ECML PKDD 2013, Prague, Czech Republic, September 23-27, 2013, Proceedings, Part III 13},
  pages={387--402},
  year={2013},
  organization={Springer}
}

@article{li2022review,
  title={A review of adversarial attack and defense for classification methods},
  author={Li, Yao and Cheng, Minhao and Hsieh, Cho-Jui and Lee, Thomas CM},
  journal={The American Statistician},
  volume={76},
  number={4},
  pages={329--345},
  year={2022},
  publisher={Taylor \& Francis}
}

@article{walke2018inception,
  title={Inception in visual cortex: in vivo-silico loops reveal most exciting images},
  author={Walke, Edgar Y and Sinz, Fabian H and Froudarakis, Emmanouil and Fahey, Paul G and Muhammad, Taliah and Ecker, Alexander S and Cobos, Erick and Reimer, Jacob and Pitkow, Xaq and Tolias, Andreas S},
  journal={bioRxiv},
  pages={506956},
  year={2018},
  publisher={Cold Spring Harbor Laboratory}
}

@inproceedings{deng2009imagenet,
  title={Imagenet: A large-scale hierarchical image database},
  author={Deng, Jia and Dong, Wei and Socher, Richard and Li, Li-Jia and Li, Kai and Fei-Fei, Li},
  booktitle={2009 IEEE conference on computer vision and pattern recognition},
  pages={248--255},
  year={2009},
  organization={Ieee}
}

@article{diffenderfer2021winning,
  title={A winning hand: Compressing deep networks can improve out-of-distribution robustness},
  author={Diffenderfer, James and Bartoldson, Brian and Chaganti, Shreya and Zhang, Jize and Kailkhura, Bhavya},
  journal={Advances in neural information processing systems},
  volume={34},
  pages={664--676},
  year={2021}
}

@inproceedings{kireev2022effectiveness,
  title={On the effectiveness of adversarial training against common corruptions},
  author={Kireev, Klim and Andriushchenko, Maksym and Flammarion, Nicolas},
  booktitle={Uncertainty in Artificial Intelligence},
  pages={1012--1021},
  year={2022},
  organization={PMLR}
}

@inproceedings{bhagoji2018enhancing,
  title={Enhancing robustness of machine learning systems via data transformations},
  author={Bhagoji, Arjun Nitin and Cullina, Daniel and Sitawarin, Chawin and Mittal, Prateek},
  booktitle={2018 52nd Annual Conference on Information Sciences and Systems (CISS)},
  pages={1--5},
  year={2018},
  organization={IEEE}
}

@inproceedings{moosavi2016deepfool,
  title={Deepfool: a simple and accurate method to fool deep neural networks},
  author={Moosavi-Dezfooli, Seyed-Mohsen and Fawzi, Alhussein and Frossard, Pascal},
  booktitle={Proceedings of the IEEE conference on computer vision and pattern recognition},
  pages={2574--2582},
  year={2016}
}

@article{paszke2017automatic,
  title={Automatic differentiation in pytorch},
  author={Paszke, Adam and Gross, Sam and Chintala, Soumith and Chanan, Gregory and Yang, Edward and DeVito, Zachary and Lin, Zeming and Desmaison, Alban and Antiga, Luca and Lerer, Adam},
  year={2017}
}

@inproceedings{modas2022prime,
  title={Prime: A few primitives can boost robustness to common corruptions},
  author={Modas, Apostolos and Rade, Rahul and Ortiz-Jim{\'e}nez, Guillermo and Moosavi-Dezfooli, Seyed-Mohsen and Frossard, Pascal},
  booktitle={European Conference on Computer Vision (ECCV)},
  pages={623--640},
  year={2022},
  organization={Springer}
}

@misc{robustness,
   title={Robustness },
   author={Logan Engstrom and Andrew Ilyas and Hadi Salman and Shibani Santurkar and Dimitris Tsipras},
   year={GitHub repository, 2019},
   url={https://github.com/MadryLab/robustness}
}

@article{carmon2019unlabeled,
  title={Unlabeled data improves adversarial robustness},
  author={Carmon, Yair and Raghunathan, Aditi and Schmidt, Ludwig and Duchi, John C and Liang, Percy S},
  journal={Advances in neural information processing systems},
  volume={32},
  year={2019}
}

@article{madry2017towards,
  title={Towards deep learning models resistant to adversarial attacks},
  author={Madry, Aleksander and Makelov, Aleksandar and Schmidt, Ludwig and Tsipras, Dimitris and Vladu, Adrian},
  journal={arXiv preprint arXiv:1706.06083},
  year={2017}
}

@inproceedings{he2016deep,
  title={Deep residual learning for image recognition},
  author={He, Kaiming and Zhang, Xiangyu and Ren, Shaoqing and Sun, Jian},
  booktitle={Proceedings of the IEEE conference on computer vision and pattern recognition},
  pages={770--778},
  year={2016}
}

@article{li2023robust,
  title={Robust deep learning object recognition models rely on low frequency information in natural images},
  author={Li, Zhe and Ortega Caro, Josue and Rusak, Evgenia and Brendel, Wieland and Bethge, Matthias and Anselmi, Fabio and Patel, Ankit B and Tolias, Andreas S and Pitkow, Xaq},
  journal={PLOS Computational Biology},
  volume={19},
  number={3},
  pages={e1010932},
  year={2023},
  publisher={Public Library of Science San Francisco, CA USA}
}

@article{brendel2017decision,
  title={Decision-based adversarial attacks: Reliable attacks against black-box machine learning models},
  author={Brendel, Wieland and Rauber, Jonas and Bethge, Matthias},
  journal={arXiv preprint arXiv:1712.04248},
  year={2017}
}

@article{rauber2017foolbox,
  title={Foolbox: A python toolbox to benchmark the robustness of machine learning models},
  author={Rauber, Jonas and Brendel, Wieland and Bethge, Matthias},
  journal={arXiv preprint arXiv:1707.04131},
  year={2017}
}

@article{sinz2018stimulus,
  title={Stimulus domain transfer in recurrent models for large scale cortical population prediction on video},
  author={Sinz, Fabian and Ecker, Alexander S and Fahey, Paul and Walker, Edgar and Cobos, Erick and Froudarakis, Emmanouil and Yatsenko, Dimitri and Pitkow, Zachary and Reimer, Jacob and Tolias, Andreas},
  journal={Advances in neural information processing systems},
  volume={31},
  year={2018}
}

@article{hendrycks2019benchmarking,
  title={Benchmarking Neural Network Robustness to Common Corruptions and Perturbations},
  author={Dan Hendrycks and Thomas Dietterich},
  journal={Proceedings of the International Conference on Learning Representations},
  year={2019}
}

@article{hinton2015distilling,
  title={Distilling the knowledge in a neural network},
  author={Hinton, Geoffrey and Vinyals, Oriol and Dean, Jeff},
  journal={arXiv preprint arXiv:1503.02531},
  year={2015}
}

@article{kriegeskorte2008representational,
  title={Representational similarity analysis-connecting the branches of systems neuroscience},
  author={Kriegeskorte, Nikolaus and Mur, Marieke and Bandettini, Peter A},
  journal={Frontiers in systems neuroscience},
  volume={2},
  pages={249},
  year={2008},
  publisher={Frontiers}
}

@article{krizhevsky2009learning,
  title={Learning multiple layers of features from tiny images},
  author={Krizhevsky, Alex and Hinton, Geoffrey and others},
  year={2009},
  publisher={Toronto, ON, Canada}
}

@article{xiao2017fashion,
  title={Fashion-mnist: a novel image dataset for benchmarking machine learning algorithms},
  author={Xiao, Han and Rasul, Kashif and Vollgraf, Roland},
  journal={arXiv preprint arXiv:1708.07747},
  year={2017}
}

@ARTICLE{726791,
  author={Lecun, Y. and Bottou, L. and Bengio, Y. and Haffner, P.},
  journal={Proceedings of the IEEE}, 
  title={Gradient-based learning applied to document recognition}, 
  year={1998},
  volume={86},
  number={11},
  pages={2278-2324},
  doi={10.1109/5.726791}}

@article{papernot2016transferability,
  title={Transferability in machine learning: from phenomena to black-box attacks using adversarial samples},
  author={Papernot, Nicolas and McDaniel, Patrick and Goodfellow, Ian},
  journal={arXiv preprint arXiv:1605.07277},
  year={2016}
}

@article{safarani2021towards,
  title={Towards robust vision by multi-task learning on monkey visual cortex},
  author={Safarani, Shahd and Nix, Arne and Willeke, Konstantin and Cadena, Santiago and Restivo, Kelli and Denfield, George and Tolias, Andreas and Sinz, Fabian},
  journal={Advances in Neural Information Processing Systems},
  volume={34},
  pages={739--751},
  year={2021}
}

@article{szegedy2013,
  title={Intriguing properties of neural networks},
  author={Szegedy, Christian and Zaremba, Wojciech and Sutskever, Ilya and Bruna, Joan and Erhan, Dumitru and Goodfellow, Ian and Fergus, Rob},
  journal={arXiv preprint arXiv:1312.6199},
  year={2013}
}

@article{goodfellow2014explaining,
  title={Explaining and harnessing adversarial examples},
  author={Goodfellow, Ian J and Shlens, Jonathon and Szegedy, Christian},
  journal={arXiv preprint arXiv:1412.6572},
  year={2014}
}

@article{croce2020robustbench,
    title={RobustBench: a standardized adversarial robustness benchmark},
    author={Croce, Francesco and Andriushchenko, Maksym and Sehwag, Vikash and Debenedetti, Edoardo and Flammarion, Nicolas
    and Chiang, Mung and Mittal, Prateek and Matthias Hein},
    journal={arXiv preprint arXiv:2010.09670},
    year={2020}
}

@inproceedings{guo2019simple,
  title={Simple black-box adversarial attacks},
  author={Guo, Chuan and Gardner, Jacob and You, Yurong and Wilson, Andrew Gordon and Weinberger, Kilian},
  booktitle={International conference on machine learning},
  pages={2484--2493},
  year={2019},
  organization={PMLR}
}
\end{small}


\newpage 

\section*{Supplementary Material}
\renewcommand{\thesection}{}

\subsection{Experimental setup}\label{experiment setup}

\noindent \textbf{Training - Neural Predictive Model.}  We train neural predictive models \cite{sinz2018stimulus, walke2018inception} to predict neural responses using all scans (measurements of neurons) with the same training configurations and neural data available in the codebase left by \cite{li2019learning}.\vspace{1ex}

\noindent \textbf{Training - Image classification.} 
All models were trained by stochastic gradient descent on a NVIDIA A100-SXM4-40GB GPU. Models classifying grayscale CIFAR-10 were trained during 40 epochs. Training and regularizing a ResNet18 CIFAR-10 took in average 34 min to run. We used a batch size of 64 for the classification pathway and a batch of 16 image pairs for the regularization pathway. The number of regularization images used by default is $5000$. Target similarities $S_{ij}^{\text{pixel}}$ are computed as follows. We compute the cosine similarity between images which are flattened, mean subtracted and normalized. We use the same learning schedule as in \cite{li2019learning}. Models were trained using Pytorch \citep{paszke2017automatic}.

The classification datasets used are : MNIST \cite{726791}, FashionMNIST \cite{xiao2017fashion}, CIFAR-10 \cite{krizhevsky2009learning},  CIFAR-100. The regularization datasets used are : MNIST, FashionMNIST,  CIFAR-10,  CIFAR-100 and  ImageNet \cite{deng2009imagenet}.  In Supplementary Material  \ref{hyperparam_val}, we report the $(\alpha, Th)$ used for each dataset combinations. The code used for training is based on the code used in \cite{li2019learning}. All codes are available at \href{https://github.com/elieattias1/pixel-reg}{https://github.com/elieattias1/pixel-reg}.\vspace{1ex}

\begin{figure}[H]
    \centering    \includegraphics[width=1.\linewidth]{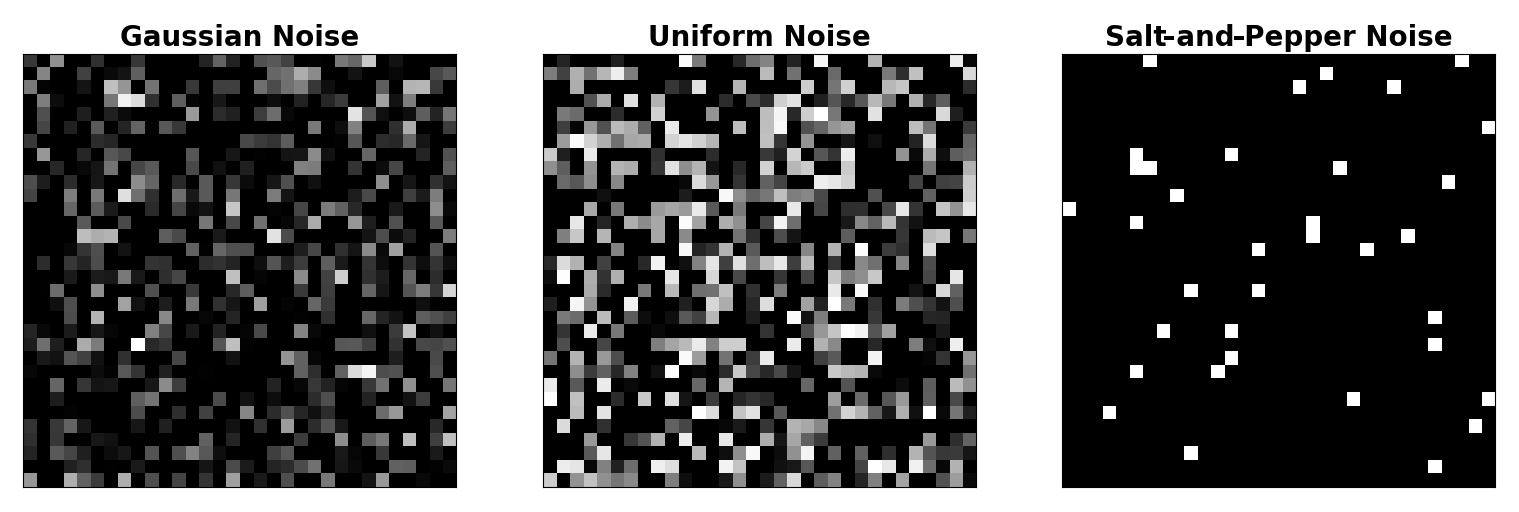}
    \caption{Visualization of (left) Gaussian noise, (center) Uniform noise and (right) Salt-and-Pepper noise with $\epsilon = 0.06$.}  \label{fig:noise_visualization}  
\end{figure}

\noindent \textbf{Adversarial attacks.} All perturbations are reported for image pixels in the range $[0, 1]$. We evaluate model robustness to random noise and transfer-based FGSM \cite{goodfellow2014explaining} perturbations by measuring the accuracy of evaluated models for distinct perturbation strengths $\epsilon$. We empirically find the range of perturbation strengths used to evaluate models, such that the unregularized model performs bad at the highest $\epsilon$ used for that particular model-dataset-attack combination. The random noise perturbations are showcased in Fig \ref{fig:noise_visualization}.

A transfer-based attack involves finding adversarial perturbations for a substitute model (an unregularized model in our case) and applying them to a target model. Evaluating robustness on transfer-based attacks is crucial because adversarial examples crafted for one model can also mislead another distinct model \cite{papernot2016transferability}.
We find these perturbations by applying the Fast Gradient Sign Method (FGSM) \cite{goodfellow2014explaining} to the substitute model. Such adversarial samples are computed as :
\begin{align}
x_{adv} = x + \epsilon \times \text{sign}(\nabla_{x}L(\theta, x, y))
\end{align}

The decision-based Boundary Attack \cite{brendel2017decision} was applied \textit{via} Foolbox \cite{rauber2017foolbox} using 50 steps, unless stated otherwise. To evaluate the success of this attack, we measure the following score: $$ S(M) = \text{median}_{i} \left(\frac{1}{d}\left\| \eta_M(x_{\text{original}}^i) \right\|_2^2 \right)$$ introduced in \cite{brendel2017decision} where $\eta_{M}(x_{original}) = x_{original} - x_{adversarial} \in \mathbb{R}^{d}$ is the adversarial perturbation found by the attack. We measure this score using $1000$ randomly sampled images from the test set. The final reported score is the average $S(M)$ calculated over 5 repetitions. All codes relative to adversarial evaluation are supplemented with this submission.

\subsection{Robustness to random noise for models trained to classify CIFAR-10 regularized using $S^{\text{pixel}}$} \label{App. pixel}

\begin{figure}[H]
    \centering    \includegraphics[width=1.\linewidth]{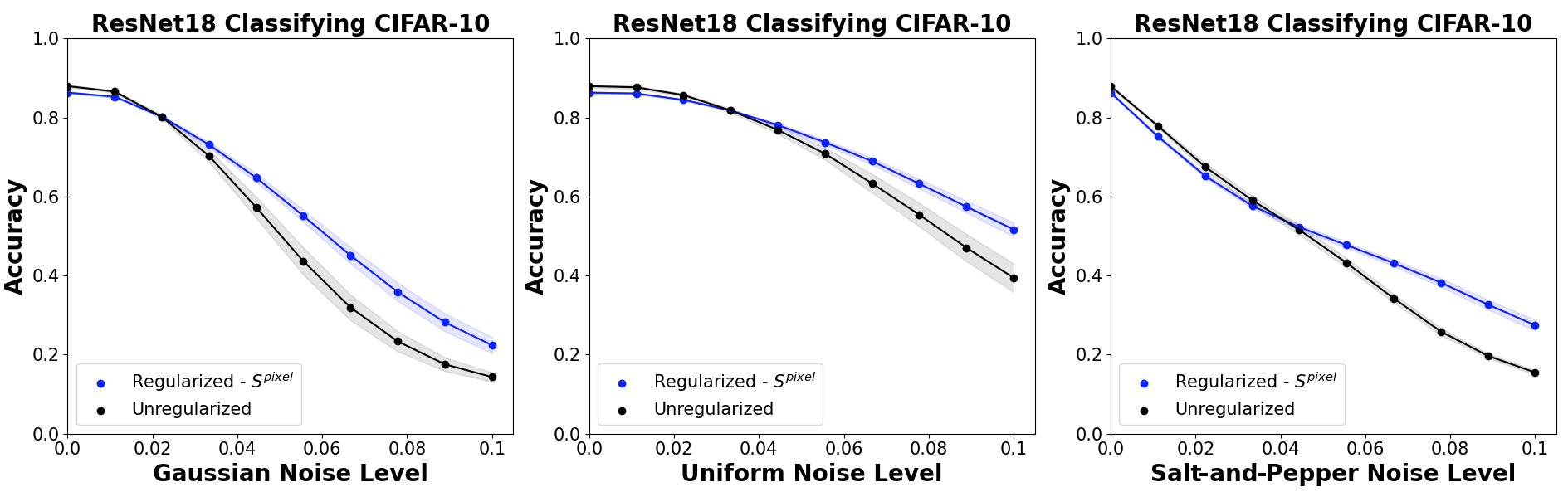}
    \caption{Robustness of a ResNet18 trained to classify grayscale CIFAR-10 and regularized on $S^{pixel}$ from grayscale images from ImageNet dataset to Gaussian, Uniform noise and Salt-and-Pepper noise. Error shades represent the SEM across seven seeds per model.}  \label{fig:App_pixel_robustness}  
\end{figure}

\subsection{Robustness to random noise for models trained to classify CIFAR-10 regularized using $S^{Th}$} \label{App. Th}
\begin{figure}[H]
    \centering
    \includegraphics[width=\linewidth]{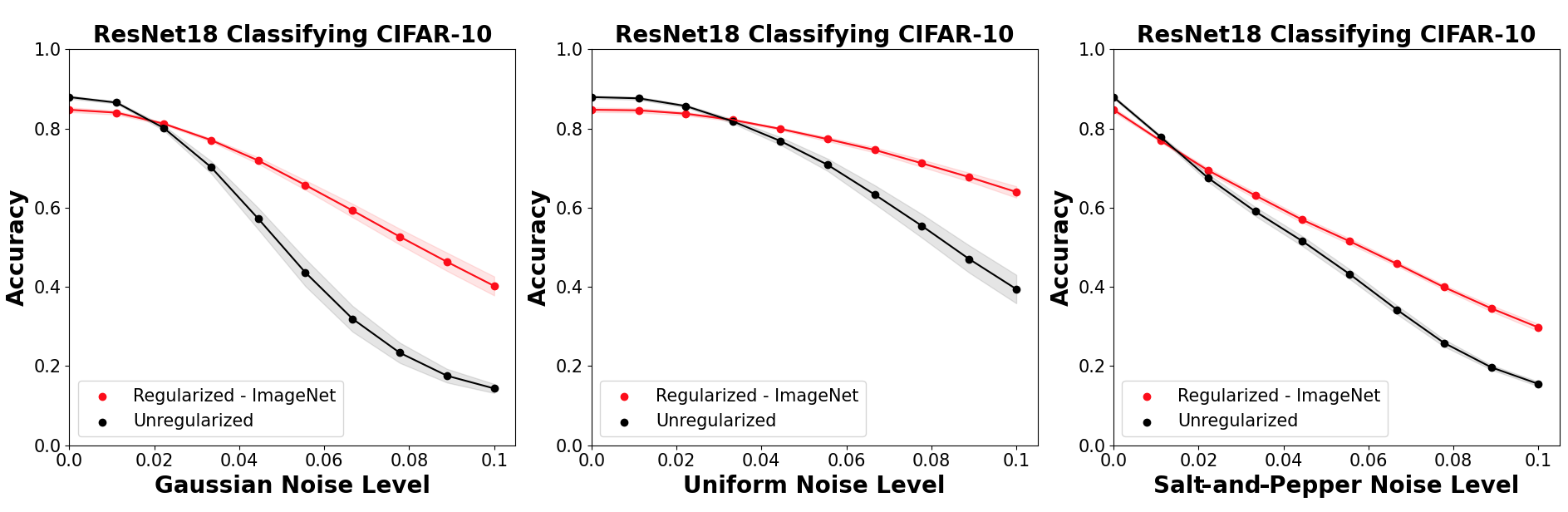}
    \caption{Robustness of a ResNet18 trained to classify grayscale CIFAR-10 and regularized on $S^{Th}$ from grayscale images from ImageNet dataset to Gaussian, Uniform noise and Salt-and-Pepper noise.  Error shades represent the SEM across seven seeds per model.}
    \label{fig:cifar10-random}
\end{figure}

\subsection{Robustness on image classification task for different classification-regularization datasets} \label{dataset_combinations}

\begin{figure}[H]
    \centering    \includegraphics[width=0.967\linewidth]{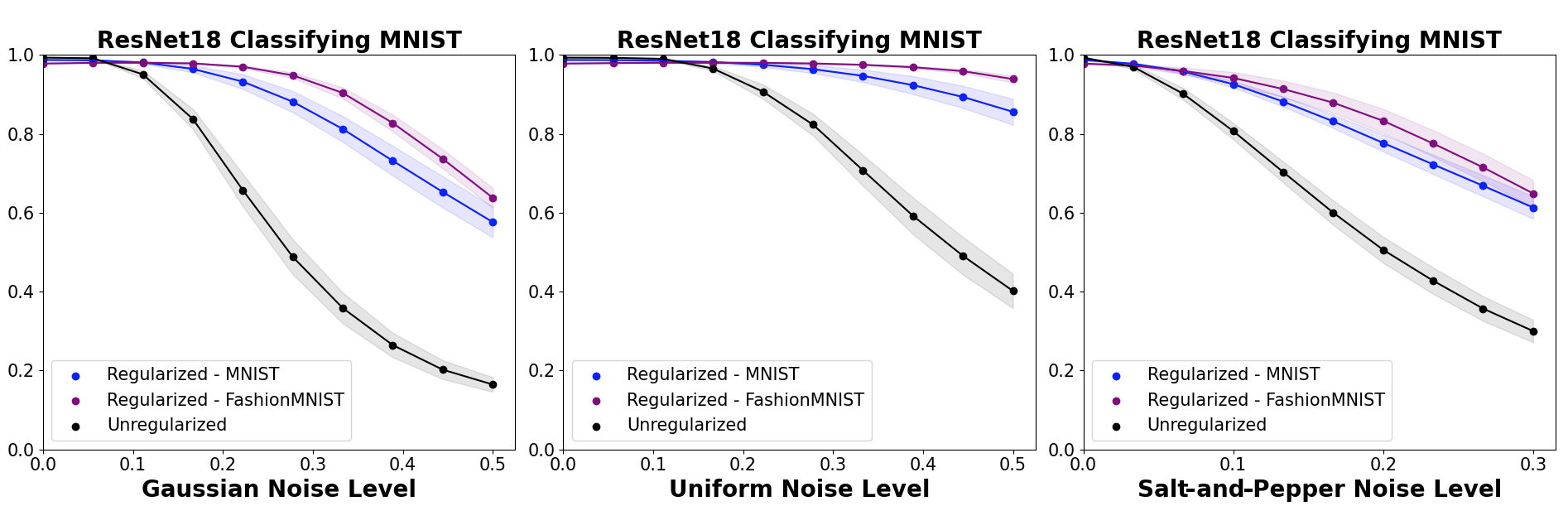}
    \caption{Robustness of a ResNet18 trained to classify MNIST and regularized on images from  MNIST and FashionMNIST datasets to Gaussian noise, Uniform noise and Salt-and-Pepper noise. Error shades/bars represent the SEM across seven seeds per model.}  \label{fig:M_robustness}  
\end{figure}

\begin{figure}[H]
    \centering    \includegraphics[width=0.967\linewidth]{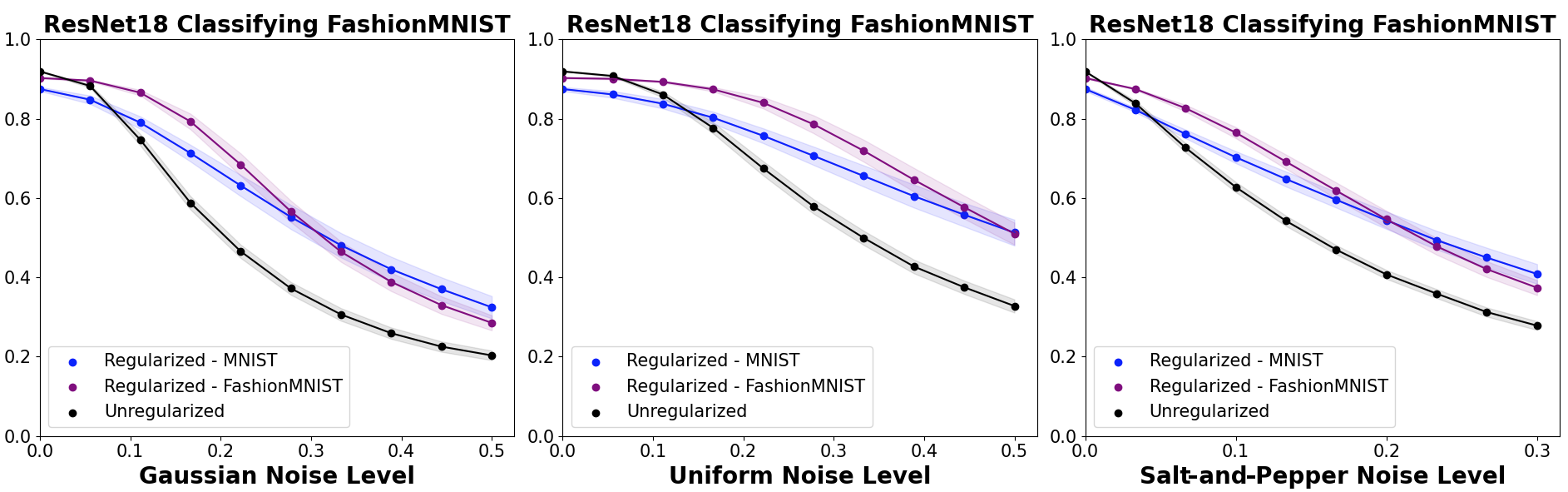}
    \caption{Robustness of a ResNet18 trained to classify FashionMNIST and regularized on images from MNIST and FashionMNIST datasets to Gaussian, Uniform and Salt-and-Pepper noise. Error shades/bars represent the SEM across seven seeds per model.}  \label{fig:F_robustness}  
\end{figure}

\begin{figure}[H]
    \centering    \includegraphics[width=0.97\linewidth]{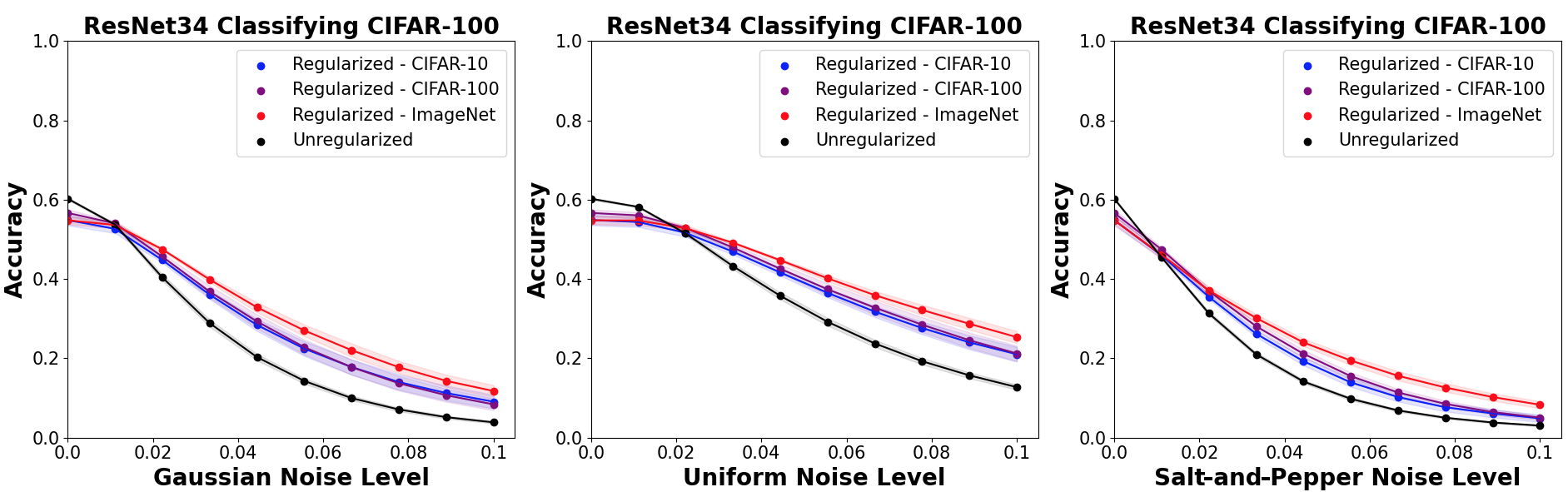}
    \caption{Robustness of a ResNet34 trained to classify grayscale CIFAR-100 and regularized on grayscale images from CIFAR-10, CIFAR-100, ImageNet datasets to Gaussian noise, Uniform noise and Salt-and-Pepper noise. Error shades/bars represent the SEM across seven seeds per model.}  \label{fig:C100_random_robustness}  
\end{figure}

\begin{figure}[H]
    \centering    \includegraphics[width=0.99\linewidth]{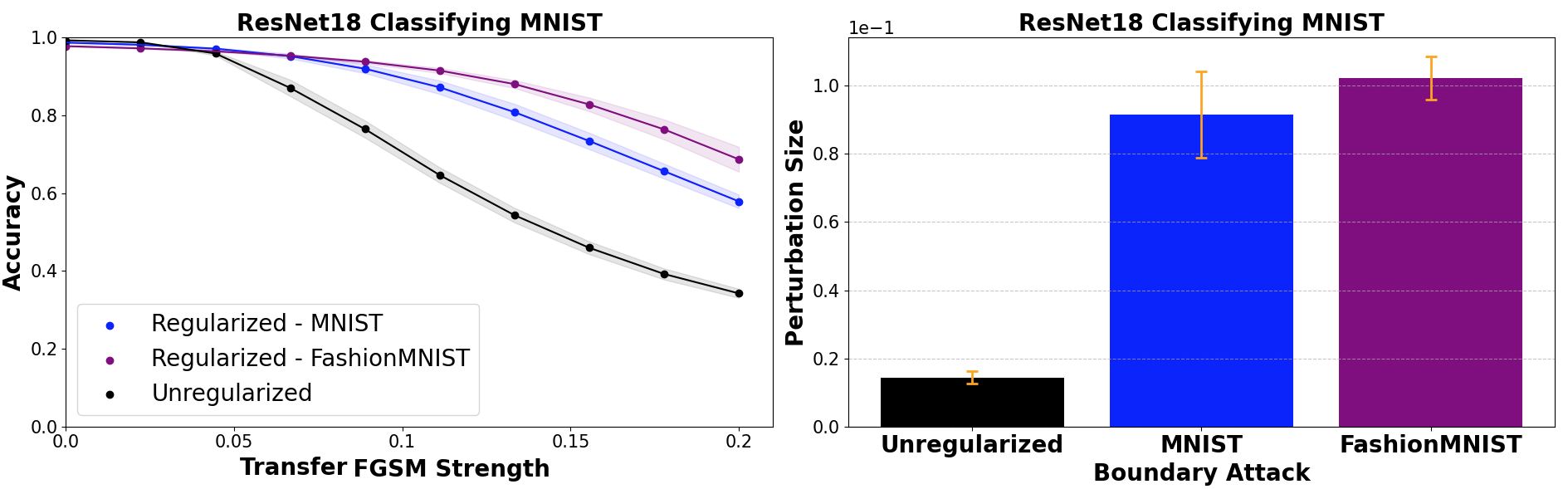}
    \caption{Robustness of a ResNet18 trained to classify MNIST and regularized on MNIST and FashionMNIST images to transfer-based FGSM \cite{goodfellow2014explaining} perturbations from an unregularized model, and a decision-based Boundary Attack \cite{brendel2017decision}. Error shades/bars represent the SEM across seven seeds per model.}  \label{fig:M_M_robustness}  
\end{figure}

\begin{figure}[H]
    \centering    \includegraphics[width=1.\linewidth]{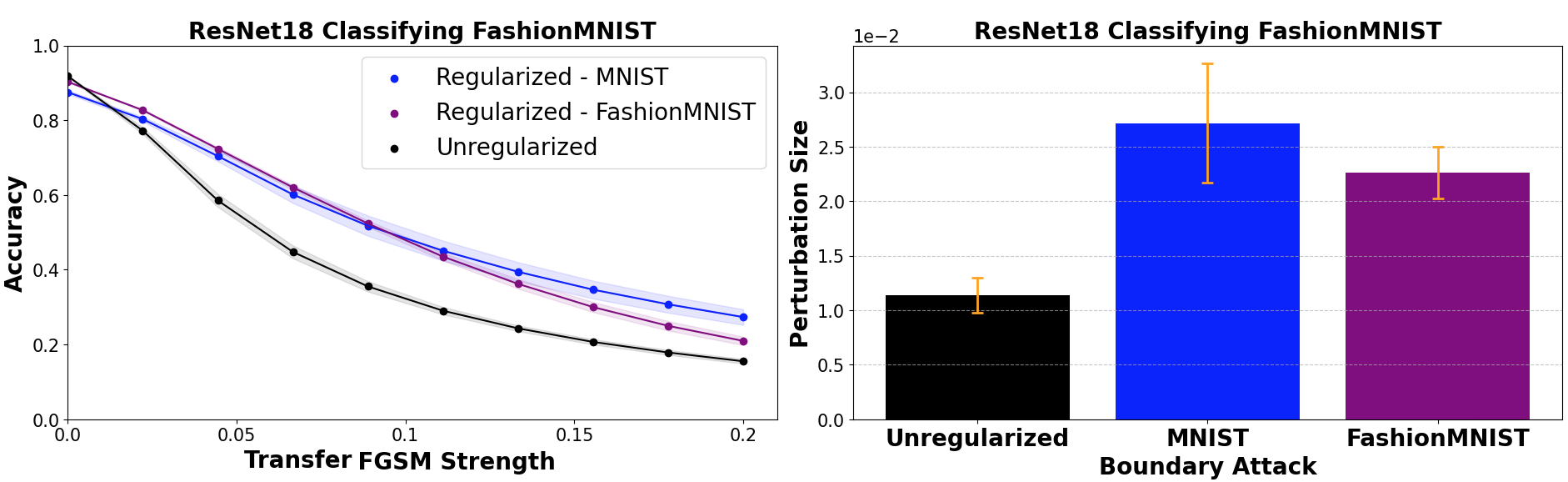}
    \caption{Robustness of a ResNet18 trained to classify FashionMNIST and regularized on images from MNIST and FashionMNIST to transfer-based FGSM \cite{goodfellow2014explaining} perturbations from an unregularized model, and a decision-based Boundary Attack \cite{brendel2017decision}. Error shades/bars represent the SEM across seven seeds per model.}  \label{fig:F_F_robustness}  
\end{figure}

\begin{figure}[H]
    \centering    \includegraphics[width=1.\linewidth]{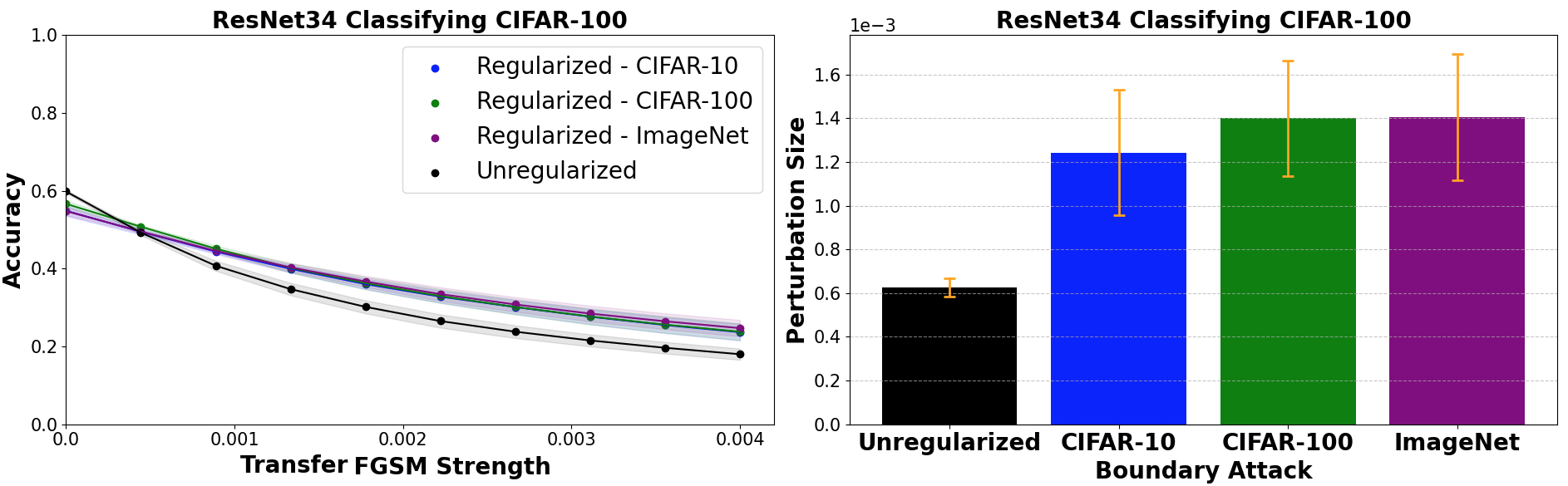}
    \caption{Robustness of a ResNet34 trained to classify grayscale CIFAR-100 and regularized on grayscale images from CIFAR-10, CIFAR-100, and ImageNet datasets to transfer-based FGSM \cite{goodfellow2014explaining} perturbations from an unregularized model, and a decision-based Boundary Attack \cite{brendel2017decision}. Error shades/bars represent the SEM across seven seeds per model.}  \label{fig:C100_IM_robustness}  
\end{figure}

\subsection{Hyperparameters used for regularization }\label{hyperparam_val}

The $\alpha, Th$ value pairs selected in our work yield an acceptable accuracy-robustness trade-off ($\frac{R_D}{U_D} > 1$ and $\frac{R_0}{U_0} \geq 0.9$) across attacks during regularization. These pairs are:

\begin{table}[H]
  \caption{Hyperparameters used for regularization}
  \label{param-table}
  \centering
  \begin{tabular}{lll}
    \toprule
    \multicolumn{2}{c}{}                   \\
     Classification - Regularization     & $\alpha$ & $Th$ \\
    \midrule
    CIFAR-10 - CIFAR-10 & 10  &0.8    \\
    CIFAR-10 - CIFAR-100 & 10 & 0.8     \\
    CIFAR-10 - ImageNet & 10      & 0.8  \\
    CIFAR-100 - CIFAR-10     & 10    & 0.8 \\
    CIFAR-100 - CIFAR-100     & 10       & 0.8  \\
    CIFAR-100 - ImageNet     & 10       & 0.8\\
    MNIST - MNIST     & 4    & 0.2  \\
    MNIST - FashionMNIST & 10       & 0.8  \\
    FashionMNIST - MNIST  & 10      & 0.4  \\
    FashionMNIST - FashionMNIST      & 10      & 0.8  \\
    \bottomrule
  \end{tabular}
\end{table}

We did not conduct an extensive hyperparameter sweep, so there are likely some $\alpha$ and $Th$ pairs that could yield stronger robustness.

\subsection{Weighting candidate layers}\label{gamma_l_values}
We here report the weights $\gamma_l$ obtained after training, for different models and dataset-combinations.

\begin{table}[h!]
  \caption{Averaged weights for all candidate layers in a ResNet18 across seven seeds per model}
  \label{res18-table}
  \centering
  \begin{tabularx}{\columnwidth}{l XXXXX}
    \toprule
     Classification - Regularization & $\gamma_1$ & $\gamma_5$ & $\gamma_9$ & $\gamma_{13}$ & $\gamma_{17}$ \\
    \midrule
    CIFAR-10 - CIFAR-10         & $0.0082$ & 0.56  & 0.38  & $0.052$ & 0 \\
    CIFAR-10 - CIFAR-100        & 0.62     & 0.38  & 0     & 0       & 0 \\
    CIFAR-10 - ImageNet         & 0.62     & 0.38  & 0     & 0       & 0 \\
    MNIST - MNIST               & 0.5      & 0.125 & 0.25  & 0.125   & 0 \\
    MNIST - FashionMNIST        & 0.125    & 0.375 & 0.5   & 0       & 0 \\
    FashionMNIST - MNIST        & 0.75     & 0.125 & 0.125 & 0       & 0 \\
    FashionMNIST - FashionMNIST & 0.143    & 0.286 & 0.571 & 0       & 0 \\
    \bottomrule
  \end{tabularx}
\end{table}

\begin{table}[h!]
  \caption{Averaged weights for all candidate layers in a ResNet34 across seven seeds per model}
  \label{res34-table}
  \centering
  \begin{tabular}{llllll}
    \toprule
    \multicolumn{2}{c}{}                   \\
     Classification - Regularization     & $\gamma_1$ & $\gamma_7 $& $\gamma_{15}$ &$ \gamma_{27}$ & $\gamma_{33}$ \\
    \midrule
    CIFAR-100 - CIFAR-10  & 0.375 & 0.375  & 0.25 & 0 & 0 \\
    CIFAR-100 - CIFAR-100 & 0.875  & 0.125  & 0  & 0& 0 \\
    CIFAR-100 - ImageNet & 0.8  & 0.2  & 0 & 0 & 0  \\
    \bottomrule
  \end{tabular}
\end{table}
\subsection{Relevance of different similarity ranges}
\label{reg_range_analysis}
In this subsection, we share an investigation regarding the target similarity ranges which matter the most for regularization. Indeed, given a threshold $Th$, specific $S_{ij}^{Th}$ will take values in $\{0, -1, 1\}$. We define the following sets of regularization target pairs : $i,j$ : $S_{-}^{Th} := \{ S_{ij} \mid  S_{ij} < -Th \text{ or }  |S_{ij}| < Th\}$ , $S_{+}^{Th} := \{ S_{ij} \mid S_{ij} >Th \text{ or }  |S_{ij}| < Th \}$, $S_{\text{low}}^{Th} :=  \{ S_{ij} \mid |S_{ij}| < Th \}$, $S_{\text{high}}^{Th} :=  \{ S_{ij} \mid |S_{ij}| > Th \}$. The set $S_{-}^{Th}$ represents regularization targets with values in $\{-1, 0\}$, while $S_{+}^{Th}$ represents targets in $\{0, 1\}$. Additionally, $S_{\text{low}}^{Th}$ represents regularization targets with values in $\{0\}$, and $S_{\text{high}}^{Th}$ represents regularization targets with values in $\{-1, 1\}$. We also define a target set depending on two thresholds : $Th_1 > Th_2>0$ denoted as $S_{\text{double}}^{Th_1, Th_2}$ such that : 
    \begin{equation}
        S_{\text{double}, ij}^{Th_1, Th_2} =  
        \begin{cases} 
             1, & \text{ if } S_{ij}^{\text{pixel}} > Th_1, \\
            -1, & \text{ if } S_{ij}^{\text{pixel}} < -Th_1, \\
             0, & \text{ if } |S_{ij}^{\text{pixel}}| \leq Th_2 
        \end{cases}.
    \end{equation}

\begin{figure}
    \centering    \includegraphics[width=1.\linewidth]{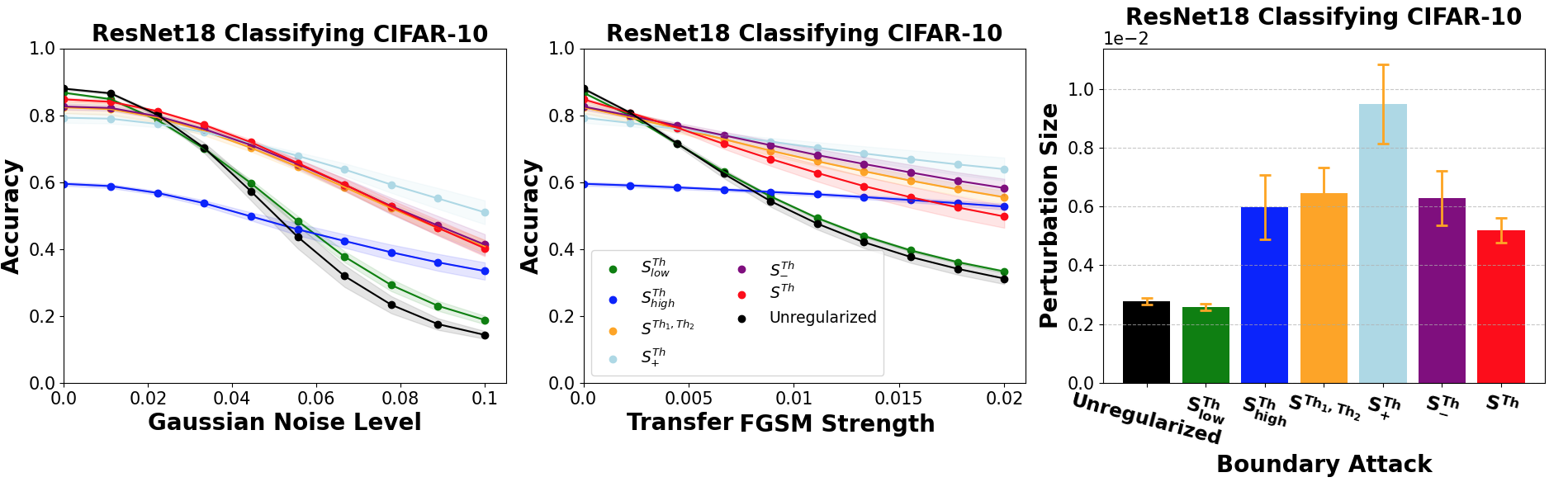}
    \caption{ResNet18 trained to classify grayscale CIFAR-10 and regularized with grayscale images from ImageNet for different regularization targets. Error shades/bars represent the SEM across seven seeds per model.}  \label{fig:C10_IM_reg_range}  
\end{figure}

\begin{figure}
    \centering    \includegraphics[width=\linewidth]{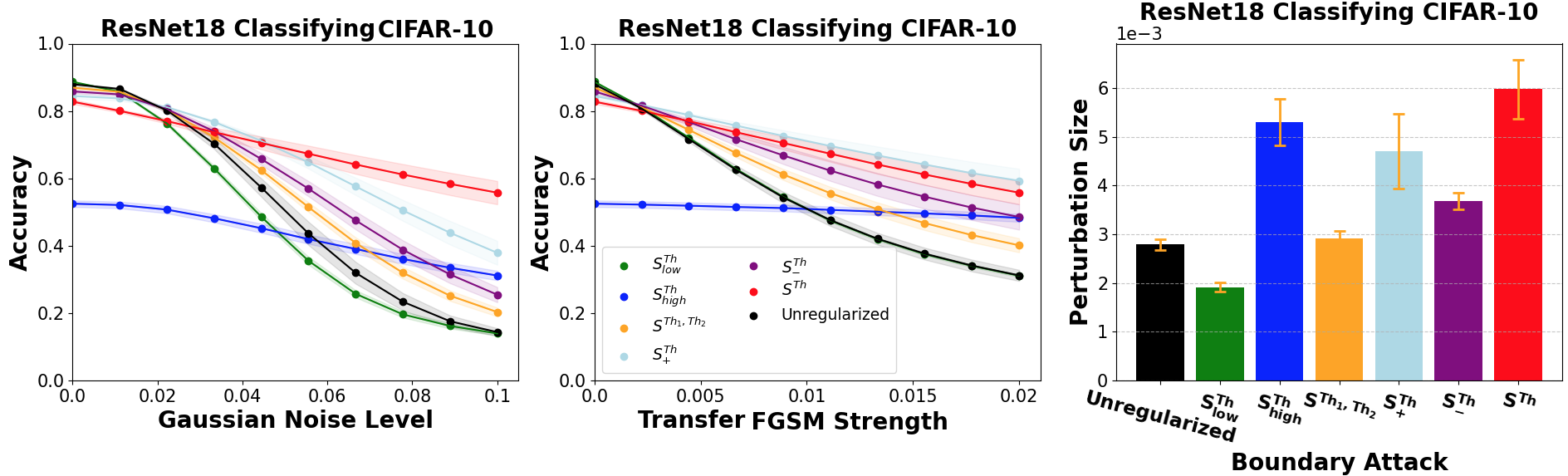}
    \caption{ResNet18 trained to classify grayscale CIFAR-10 and regularized with grayscale images from CIFAR-10 for different regularization targets. Error shades/bars represent the SEM across seven seeds per model. }  \label{fig:C10_C10_reg_range}  
\end{figure}

\begin{figure}[H]
    \centering    \includegraphics[width=\linewidth]{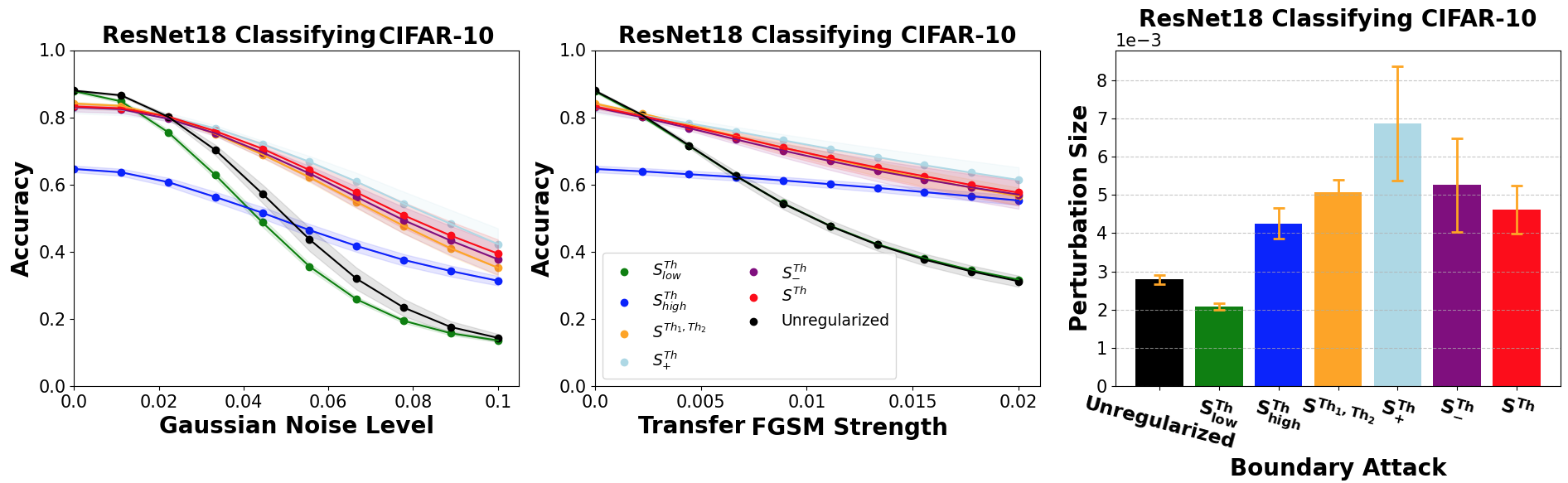}
    \caption{ResNet18 trained to classify grayscale CIFAR-10 and regularized with grayscale images from CIFAR-100 for different regularization targets. Error shades/bars represent the SEM across seven seeds per model.}  \label{fig:C10_C100_reg_range}  
\end{figure}

\subsection{Regularization on color datasets}\label{App - colored CIFAR100}
Here, we show that our method is also successful when using color datasets, which are more utilized in practice. 

\begin{figure}[H]
    \centering    \includegraphics[width=0.95\linewidth]{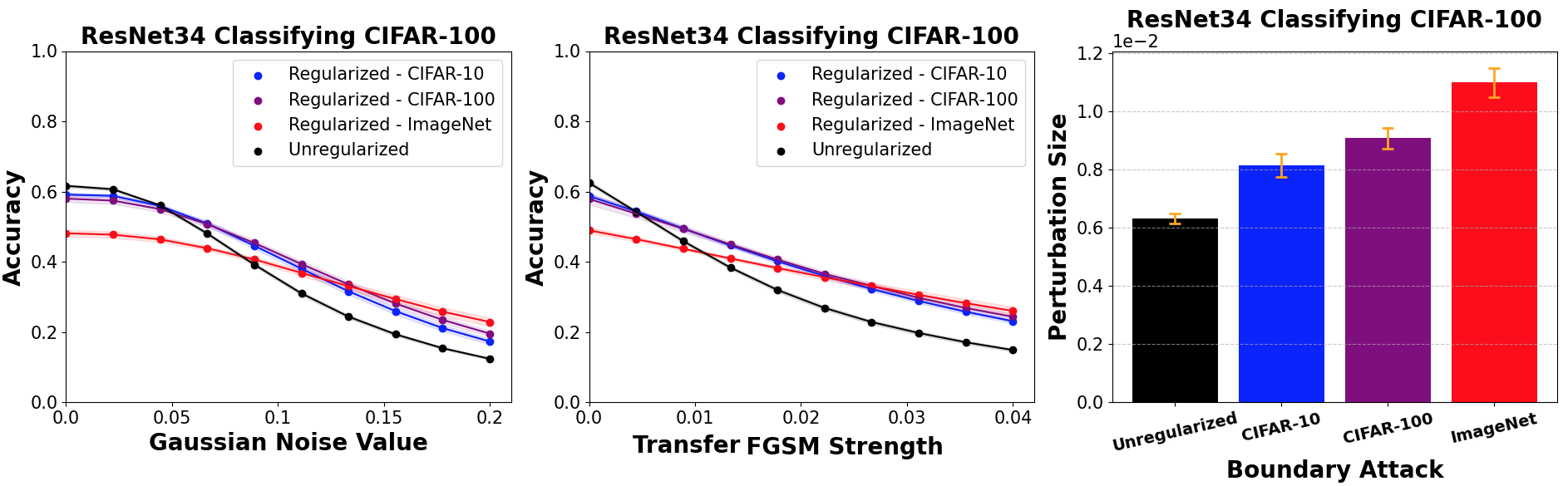}
    \caption{Robustness of a ResNet34 trained to classify color CIFAR-100 regularized on color images from different datasets : CIFAR-10 (blue), CIFAR-100 (purple) or ImageNet (red). For the decision-based Boundary Attack, we compute the median $L_2$ perturbation size, averaged across 1000 images, and 5 repeats. Error shades/bars represent the SEM across seven seeds per model. }
    \label{fig:color-cifar100}
\end{figure}

In Fig. \ref{fig:color-cifar100} we show results using color CIFAR-100 as classification dataset, and color CIFAR-10, CIFAR-100 or ImageNet as regularization datasets. As seen, there is an increase in the model's robustness for all regularization datasets.

\ignore{
\subsection{Benchmarking our method against those in the RobustBench leaderboard}

We benchmarked a model trained to classify (color) CIFAR-10 regularized on (color) CIFAR-100 using our method, against 3 models referenced in the RobustBench \cite{hendrycks2019robustness} leaderboard - commonly used to systematically track the real progress in adversarial robustness - on CIFAR-10-C, at severity 5. We select $L_{\infty}$, $L_2$ and common corruption (thereafter CC) specific models denoted as Carmon2019Unlabeled \cite{carmon2019unlabeled}, Engstrom2019Robustness \cite{robustness} , Modas2021PRIMEResNet18 \cite{modas2022prime} respectively. We also benchmark against the 'standard' model available on RobustBench. As we see in Fig. \ref{fig:benchmarking}, our regularized model performs better than the standard model reflecting the robustness gain, but does not reach the same performance as state of the art networks. We acknowledge that our method does not beat the state of the art methods in the adversarial robustness literature. The aim of our method is not to beat the state of the art methods, but rather to show that our method which is based on the neural regularizer in \cite{li2019learning} can be equally effective without the need to use expensive neural data.

\begin{figure}[H]
    \centering    \includegraphics[width=0.45\linewidth]{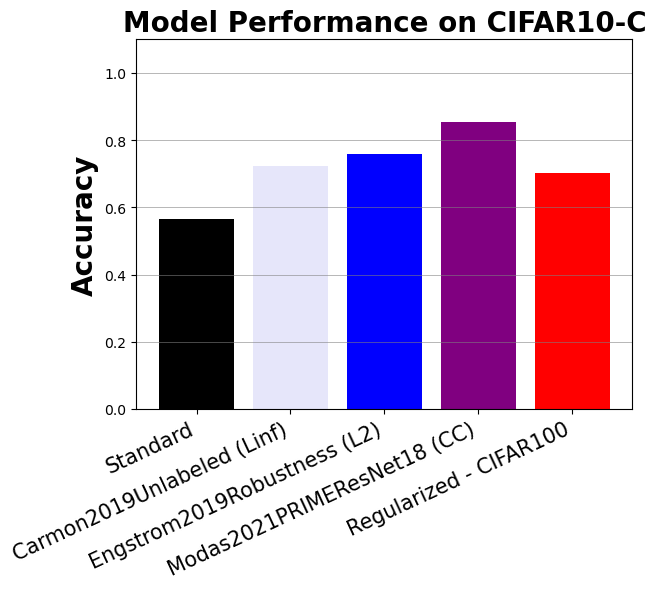}
    \caption{Benchmarking a model trained to classify (color) CIFAR-10 regularized on (color) CIFAR-100 using our method, against 3 models referenced in the RobustBench \cite{croce2020robustbench} leaderboard - commonly used to systematically track the real progress in adversarial robustness - on CIFAR-10-C, at severity 5. We select $L_{\infty}$, $L_2$ and common corruption (thereafter CC) specific models denoted as Carmon2019Unlabeled \cite{carmon2019unlabeled}, Engstrom2019Robustness \cite{robustness} , Modas2021PRIMEResNet18 \cite{modas2022prime} respectively. We selected our most robust model on CIFAR10-C for such benchmarking. We also benchmark against the 'standard' model available on RobustBench.} 
    \label{fig:benchmarking}
\end{figure}

}

\subsection{Fourier spectra of common corruptions}
Following the approach of \cite{li2023robust}, we divided the common corruptions present in CIFAR-10-C into three categories based on their corresponding dominating frequencies whether they belong to low, medium or high frequency ranges. Low frequency corruptions are composed of ‘snow’, ‘frost’, ‘fog’, ‘brightness’, ‘contrast’ corruptions. Medium-frequency corruptions are composed of ‘motion blur’, ‘zoom blur’, ‘defocus blur’, ‘glass blur’, ‘elastic transform’, ‘jpeg compression’ and ‘pixelate’ corruptions. High-frequency corruptions are composed of ‘gaussian noise’, ‘shot noise’ and ‘impulse noise’ corruptions. For completeness, we reproduced the Fourier transformation of the common corruptions as shown below.

\begin{figure}[H]
    \centering    \includegraphics[width=0.9\linewidth]{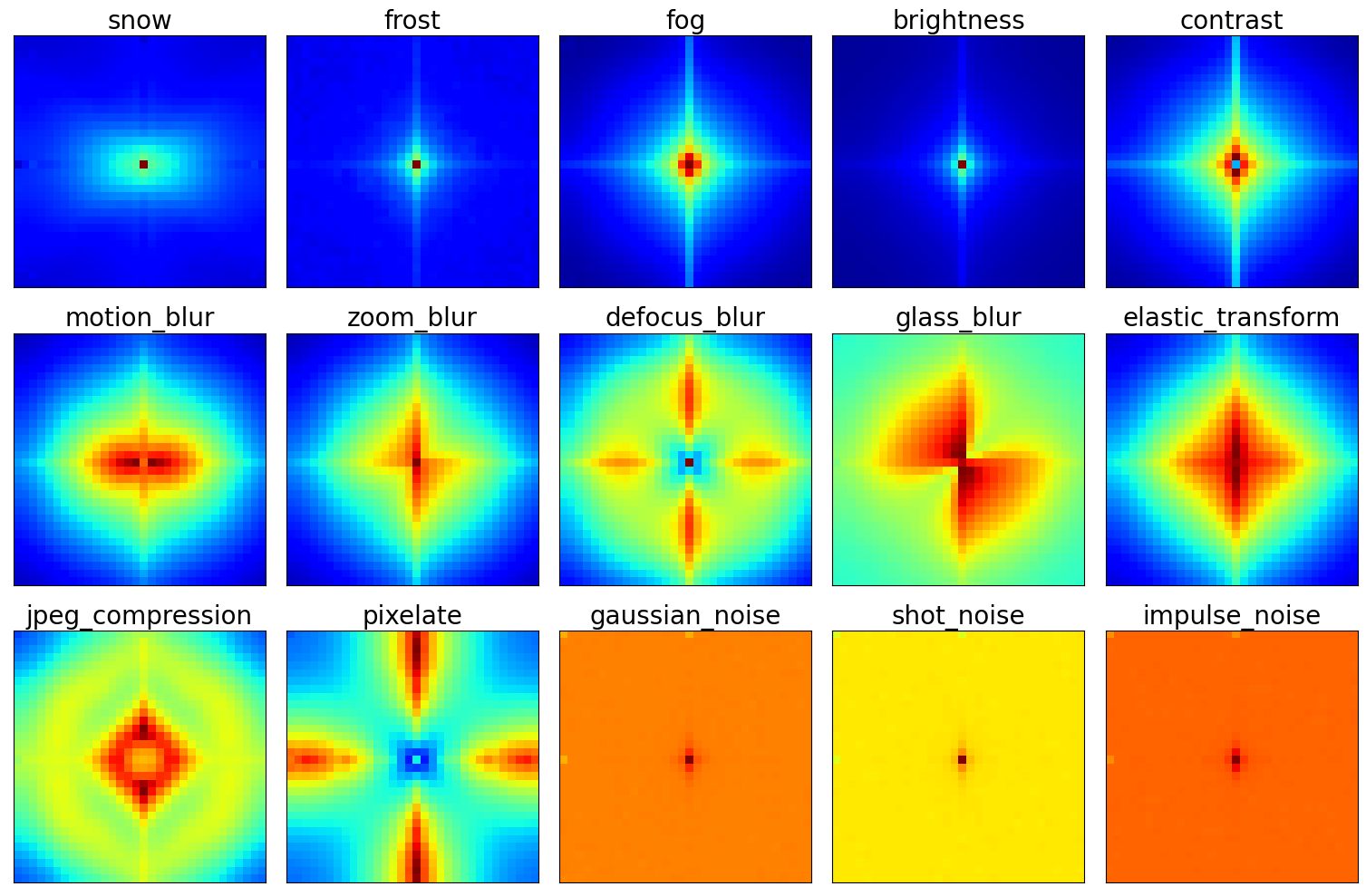}
    \caption{Fourier transform of the corruptions in CIFAR-10-C at severity 3, ordered by groups of frequency ranges from low, to medium to high. For each corruption, we compute $\mathbb{E}[|F(C(X) - X)|]$ by averaging over all test images, where $F$ denotes a discrete Fourier transformation, $X$ denotes an original image, and $C(X)$ its corrupted counterpart. We apply a logarithm $x\mapsto \log (1 + x)$ for visualization purposes.}  \label{fig:CC_spectra}  
\end{figure}

\end{document}